%% file: paper.tex
\pdfoutput=1
\documentclass[conference]{IEEEtran}
\usepackage{amsmath,amssymb}
\usepackage{graphicx,booktabs,xcolor,tikz}
\usetikzlibrary{arrows.meta,positioning,fit,backgrounds,calc}
\definecolor{ours}{HTML}{D1495B}\definecolor{chip}{HTML}{2E4057}\definecolor{mute}{HTML}{5C677D}
\usepackage[colorlinks,linkcolor=black,citecolor=black,urlcolor={rgb:red,0;green,0.27;blue,0.73}]{hyperref}
\usepackage[capitalise,noabbrev]{cleveref}
\graphicspath{{figures/}}
\input{tables/chipfid.tex}
\newcommand{\hailo}{Hailo-8L}
\newcommand{\rpi}{Raspberry Pi~5}
\newcommand{\dav}{Depth Anything~V2}
\newcommand{\keyline}[1]{\smallskip\noindent\textbf{#1}\ }
\begin{document}
\title{Seeing to Drive on a Single Board:\\ Attention-Free Depth for Onboard Robot Navigation}
\author{\IEEEauthorblockN{Itai Mizlish and Boaz Ben-Moshe}
\IEEEauthorblockA{Department of Computer Science, Ariel University, Ariel, Israel\\
itai.mizlish@msmail.ariel.ac.il, benmo@g.ariel.ac.il}}
\maketitle

\begin{abstract}
A small ground robot can carry one camera and a few watts of compute, but the depth foundation models that could turn
that camera into a range sensor are vision transformers sized for GPUs. We bring one to a \rpi{} with the \hailo{}, an
accelerator that runs every layer in integer arithmetic, attention included. Quantised for it, \dav{} loses large parts of
the scene, and we measure why: at 449 tokens, 91\,\% of softmax weights fall below one int8 step while carrying
17--33\,\% of the attention mass. Our answer is to distil, not quantise. An attention-free ResNet-34 U-Net student of
\dav-Large (24.5\,M parameters) runs at 27.1~FPS on the chip, its int8 output matches its float twin (Pearson
\chipfidvfivepfifty), and on held-out scenes of its simulated yard it is the most accurate depth network on the chip
(AbsRel 0.269 against 0.354--0.530). On 1{,}000 stereo-verified points of a real outdoor drive it is not resolvably
different from the vendor's \dav-Small build, while running faster (27.1 vs 21.6~FPS). Ranging obstacles by where they
meet the floor needs only the camera height, so mapping, hybrid-A$^\ast$ planning and SLAM run on the Pi's own CPU. In
real-time hardware-in-the-loop tests in a Gaussian-splat reconstruction of a real yard, the stack reaches the goal without
contact in 7.9$\pm$1.1 of 10 scenes, against 9.2$\pm$0.4 with exact obstacles, and in 7.7$\pm$0.7 on its own pose with
simulated wheel odometry (ten seeds each; 8.2$\pm$0.4 with the real Pi and \hailo{} in the loop, five seeds). The source
code, pre-trained models and data are available at \url{https://github.com/itaim18/DANIEL}.
\end{abstract}

\begin{figure*}[t]
\centering
\includegraphics[width=\textwidth]{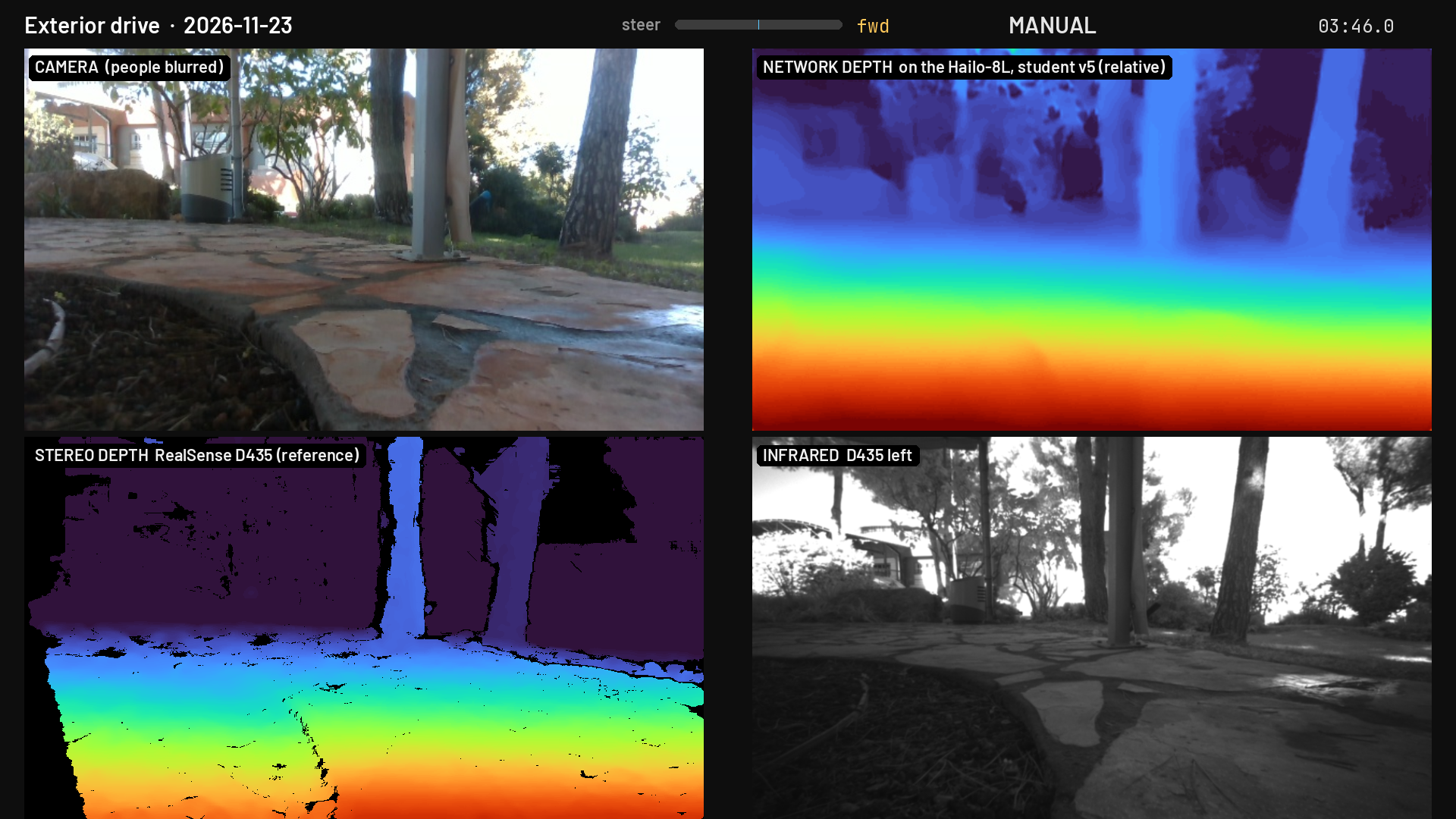}
\caption{What the car sees: one moment of an exterior drive on the real robot (hand-driven, November 2026). Top left, the
robot's RGB camera (people blurred); top right, relative depth from our attention-free student, computed onboard on the
\hailo{}; bottom left, RealSense D435 stereo depth, recorded only as a reference; bottom right, the D435 left infrared
image. Warm colours are near. The header gives the session clock, the steering and throttle sent to the car, and the
driving mode. The pole and the pine trunks ahead of the car show up in all four views.}
\label{fig:drive_views}
\end{figure*}

\section{Introduction}
\begin{figure*}[t]
\centering
\begin{tikzpicture}[lab/.style={font=\sffamily\footnotesize, align=left, inner sep=1pt},
  lead/.style={thick, ours!85!black}, dot/.style={circle, fill=ours, draw=white, thick, inner sep=1.6pt}]
\node[anchor=south west, inner sep=0] (f) at (0,0) {\includegraphics[width=0.58\textwidth]{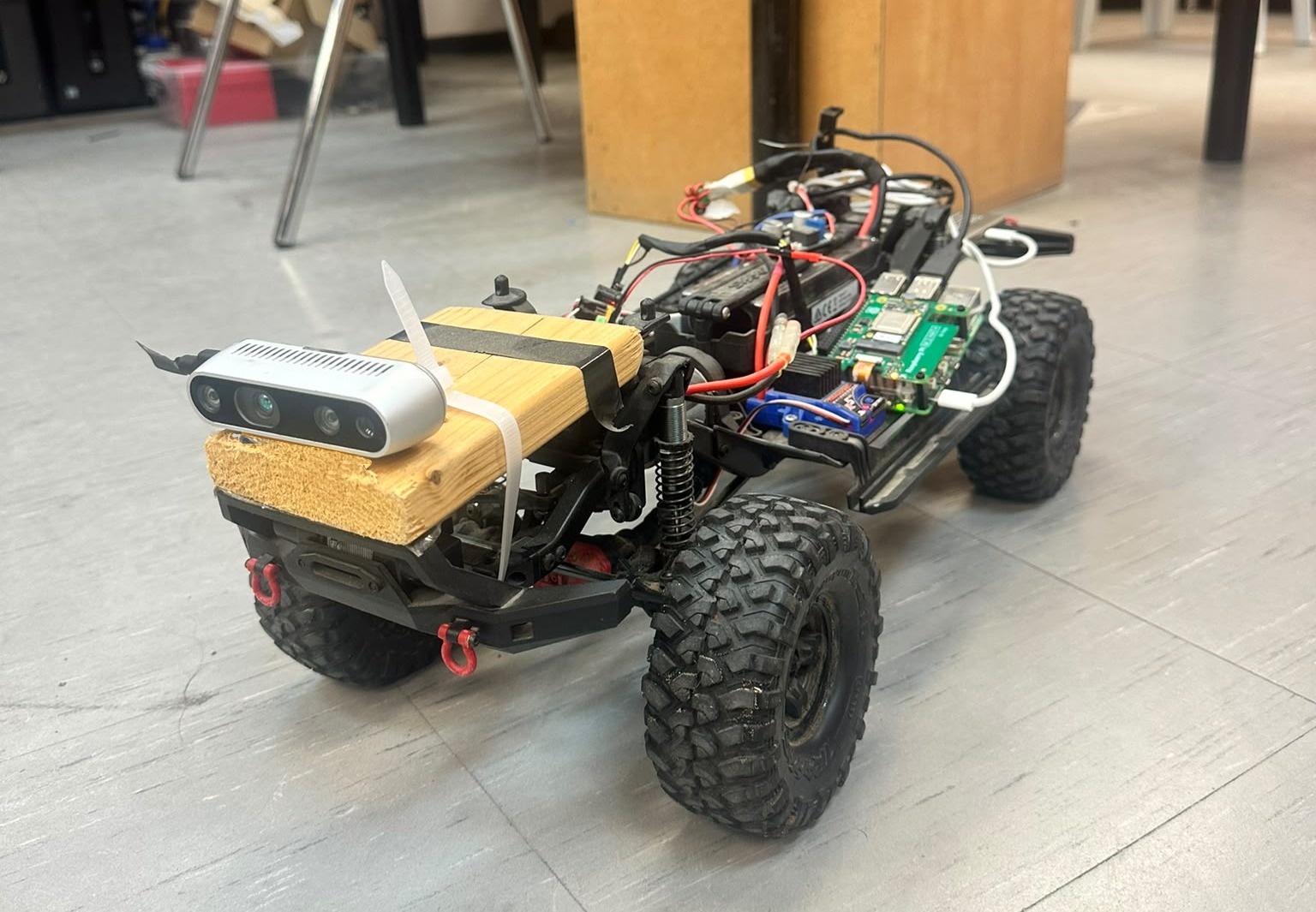}};
\begin{scope}[x={(f.south east)}, y={(f.north west)}]
\node[lab, anchor=east, align=right] (cam) at (-0.03,0.80) {\textbf{Intel RealSense D435}\\camera; colour\\image only};
\node[lab, anchor=west] (pca) at (1.03,0.90) {\textbf{PCA9685 PWM driver}\\steering servo (ch.~0)\\and ESC (ch.~3)};
\node[lab, anchor=west] (pi) at (1.03,0.62) {\textbf{Raspberry Pi~5 +}\\\textbf{Hailo-8L AI HAT}\\runs everything in\\\cref{fig:system}};
\node[lab, anchor=west] (esc) at (1.03,0.34) {\textbf{Electronic speed}\\\textbf{controller (ESC)}\\drives the motor};
\draw[lead] (cam.east) -- (0.24,0.56) node[dot]{};
\draw[lead] (pca.west) -- (0.609,0.75) node[dot]{};
\draw[lead] (pi.west) -- (0.684,0.628) node[dot]{};
\draw[lead] (esc.west) -- (0.635,0.548) node[dot]{};
\end{scope}
\end{tikzpicture}
\caption{The robot: a Traxxas rock crawler with a 0.56$\times$0.29\,m footprint. The onboard system uses only the colour
image of the front camera; the camera's stereo depth serves as a reference in \cref{sec:real}. The \rpi{} with the \hailo{} AI
HAT runs the whole system of \cref{fig:system} and commands the steering servo and the electronic speed controller (ESC)
through the PCA9685 PWM driver.}
\label{fig:robot}
\end{figure*}

A rover that drives through a yard of chairs, cones, crates and people must know where things are. Range sensors answer
that at a cost in money, power, mass and blind spots: a planar LiDAR misses a table top and a low curb, and a stereo
camera adds a baseline, power and cost (ours carries one only as a reference, \cref{fig:robot}). A single RGB camera sees all
of these, costs little and emits nothing, but gives no distance directly.

Monocular depth has changed character in the last three years. Networks trained on tens of millions of images, from
MiDaS~\cite{ranftl2022midas} to Depth Anything~\cite{yang2024da1}, \dav{}~\cite{yang2024dav2}, Depth
Pro~\cite{bochkovskiy2025depthpro} and Depth Anything~3~\cite{lin2025da3}, now predict relative depth sharp and stable enough
to drive with. They are vision transformers sized for data-centre GPUs: \dav-Large has 335\,M parameters and takes 175\,ms per
frame even on the RTX~5090 laptop GPU used in this work. A small robot has a few watts, and none of these models targets an
integer-only accelerator.

We ask how much of that capability survives on the cheapest serious edge platform available today, a \rpi{} with the
13-TOPS \hailo{} of the Raspberry Pi AI Kit~\cite{rpi2024aikit}, and what a rover can do with it. The answer is not a
compilation exercise. The accelerator executes every layer in integer arithmetic, attention included, and both our int8
build of \dav-Small and the vendor's lose large parts of the scene. Even a faithful depth map is only relative, so it
cannot by itself tell a planner that a chair is 1.2\,m away. And a car needs a map, a planner, a controller and an estimate
of its own position within the same board and control period.

\keyline{Contributions.}
\begin{enumerate}
  \item A measured mechanism for why transformer depth fails on the \hailo{} (int8 cannot represent diffuse attention
  at a few hundred tokens), and an attention-free student distilled from \dav-Large whose int8 output equals its float
  output on the chip (\cref{sec:int8,sec:student}).
  \item A head-to-head benchmark of every depth network we could run on the \hailo{}, on 400 simulated frames from held-out
  scenes with metric ground truth and 100 real robot frames: on the simulated frames the student is the most accurate on the chip, and it
  is faster than both \dav-Small builds (\cref{sec:bench}). On 1{,}000 stereo-verified points of a real outdoor drive it is
  not resolvably different from the vendor's \dav-Small (\cref{sec:real}).
  \item Scale-free metric ranging from relative depth by ground-contact geometry, and a complete onboard stack on the Pi
  (map, hybrid A$^\ast$, pure pursuit, ground-plane SLAM with wheel odometry and a slip guard) with its measured real-time
  budget (\cref{sec:space,sec:autonomy}).
  \item A real-time hardware-in-the-loop evaluation with the real board, an explicit error budget from oracle to
  camera-only odometry, and the negative results reported with the same rigour as the positive ones (\cref{sec:hitl}).
\end{enumerate}

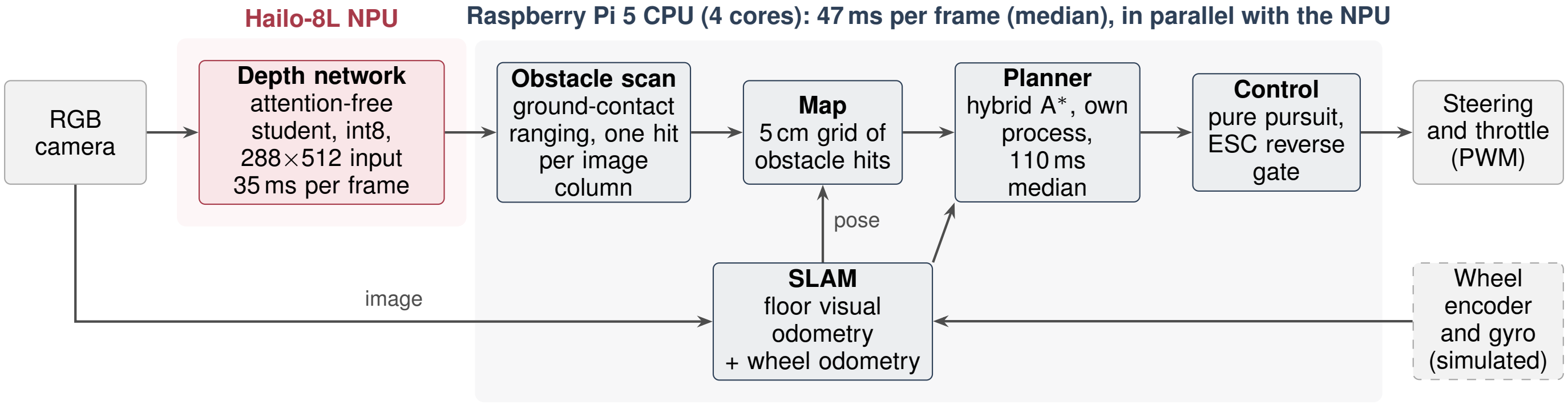
\begin{figure*}[t]
\centering
\begin{tikzpicture}[font=\sffamily\footnotesize, node distance=4mm and 6mm,
  box/.style={draw, rounded corners=2pt, align=center, minimum height=12mm, text width=#1, inner sep=2pt},
  npu/.style={box=#1, fill=ours!14, draw=ours!80!black}, npu/.default=27mm,
  cpu/.style={box=#1, fill=chip!9, draw=chip}, cpu/.default=20mm,
  io/.style={box=#1, fill=gray!10, draw=gray!70}, io/.default=15mm,
  arr/.style={-{Stealth[length=1.8mm]}, thick, gray!60!black}]
\node[io] (cam) {RGB\\camera};
\node[npu, right=of cam] (npu) {\textbf{Depth network}\\attention-free\\student, int8,\\288$\times$512 input\\35\,ms per frame};
\node[cpu=21mm, right=of npu] (scan) {\textbf{Obstacle scan}\\ground-contact\\ranging, one hit\\per image column};
\node[cpu=17mm, right=of scan] (map) {\textbf{Map}\\5\,cm grid of\\obstacle hits};
\node[cpu=20mm, right=of map] (plan) {\textbf{Planner}\\hybrid A$^\ast$, own\\process, 110\,ms\\median};
\node[cpu=18mm, right=of plan] (ctl) {\textbf{Control}\\pure pursuit,\\ESC reverse\\gate};
\node[io=16mm, right=of ctl] (pwm) {Steering\\and throttle\\(PWM)};
\node[cpu=24mm, below=9mm of map] (slam) {\textbf{SLAM}\\floor visual odometry\\+ wheel odometry};
\node[io=16mm, dashed, below=9mm of pwm] (odo) {Wheel\\encoder\\and gyro\\(simulated)};
\draw[arr] (cam) -- (npu); \draw[arr] (npu) -- (scan);
\draw[arr] (scan) -- (map); \draw[arr] (map) -- (plan); \draw[arr] (plan) -- (ctl);
\draw[arr] (ctl) -- (pwm); \draw[arr] (odo.west) -- (odo.west -| slam.east);
\draw[arr] (cam.south) |- node[pos=0.75, above, font=\sffamily\scriptsize]{image} (slam.west);
\draw[arr] (slam.north) -- node[right, font=\sffamily\scriptsize]{pose} (map.south);
\draw[arr] (slam.north east) -- (plan.south west);
\begin{scope}[on background layer]
\node[fit=(npu), inner sep=2.5mm, fill=ours!5, rounded corners=3pt, label={[font=\sffamily\footnotesize\bfseries, text=ours!80!black]above:Hailo-8L NPU}] {};
\node[fit=(scan) (map) (plan) (ctl) (slam), inner sep=2.5mm, fill=chip!4, rounded corners=3pt,
  label={[font=\sffamily\footnotesize\bfseries, text=chip]above:Raspberry Pi~5 CPU (4 cores): 47\,ms per frame (median), in parallel with the NPU}] {};
\end{scope}
\end{tikzpicture}
\caption{The onboard system. Red: runs on the \hailo{} NPU; blue: runs on the \rpi{}'s CPU; grey: sensors and actuators; dashed: simulated in the evaluation. The
NPU computes depth for the next frame while the CPU processes the current one (\cref{fig:timing}), and the planner runs in its
own process, so a slow search never stalls the frame loop. Metric range comes from the camera height, not from the depth
network (\cref{sec:space}).}
\label{fig:system}
\end{figure*}

\Cref{fig:drive_views} shows what the car sees on an outdoor drive, \cref{fig:robot} the robot, and \cref{fig:system} the
complete onboard system and where each part runs.

\section{Related Work}
\keyline{Monocular depth.} MiDaS~\cite{ranftl2022midas} introduced the scale- and shift-invariant loss that makes relative
depth transfer zero-shot, DPT~\cite{ranftl2021dpt} moved it to a vision transformer, and Depth Anything~\cite{yang2024da1}
scaled pseudo-labelling to 62\,M images. \dav{}~\cite{yang2024dav2} trains a large teacher on synthetic ground truth and
distils it into 25--335\,M-parameter students. Metric models condition on or estimate the camera~\cite{hu2024metric3dv2,
piccinelli2024unidepth,bochkovskiy2025depthpro}; Depth Anything~3~\cite{lin2025da3} predicts consistent geometry from any
number of views; Distill Any Depth~\cite{he2025distillanydepth} shows that distillation choices alone strengthen a student.
All are transformers, and none targets an integer-only accelerator.

\keyline{Depth on small computers.} FastDepth~\cite{wofk2019fastdepth} runs a pruned MobileNet encoder-decoder on a Jetson;
Lite-Mono~\cite{zhang2023litemono} and SC-DepthV3~\cite{sun2024scdepthv3} learn from video without labels. FastDepth and
SC-DepthV3 are the two depth models Hailo ships for the \hailo{}, and both appear in our benchmark. AsyncMDE~\cite{ma2026asyncmde}
amortises a foundation model over time on a Jetson AGX Orin. Vision transformers are hard to quantise because post-softmax
activations are long-tailed; FQ-ViT~\cite{lin2022fqvit}, PTQ4ViT~\cite{yuan2022ptq4vit} and RepQ-ViT~\cite{li2023repqvit}
need quantiser support that a fixed NPU toolchain may not offer.

\keyline{Navigating with predicted depth.} MonoNav~\cite{simon2023mononav} fuses transformer depth into a reconstruction to
fly a Crazyflie, with depth computed off-board. MonoMPC~\cite{sharma2026monompc} uses depth as context for a learned
collision model. Finke et al.~\cite{finke2026omninav} fine-tune \dav{} for four cameras and run an RL policy on a Jetson AGX
Orin at 10\,Hz. Learned policies~\cite{shah2023vint,sridhar2024nomad} and imperative planners~\cite{yang2023iplanner,roth2024viplanner}
map images to paths. We keep planning classical (hybrid A$^\ast$~\cite{dolgov2010hybrid}, pure
pursuit~\cite{coulter1992purepursuit}) so that every failure can be attributed to perception, planning or localisation, and
run all of it on a board an order of magnitude cheaper than an AGX Orin.

\section{Why Transformer Depth Breaks on the Hailo-8L}\label{sec:int8}
The \hailo{} runs every layer with 8-bit weights and activations, including the matrix products and softmax of attention.
We exported \dav-Small at 224$\times$392 (a 16$\times$28 grid of 14-pixel patches plus a class token: 449 tokens),
calibrated it on domain images and compiled it. On real robot video the result looked right (median Pearson 0.99 to the
float model). On the yard renders it erased obstacles on a third of held-out frames, dissolving a crate or a person into the
floor.

\begin{figure*}[t]
\centering
\includegraphics[width=\textwidth]{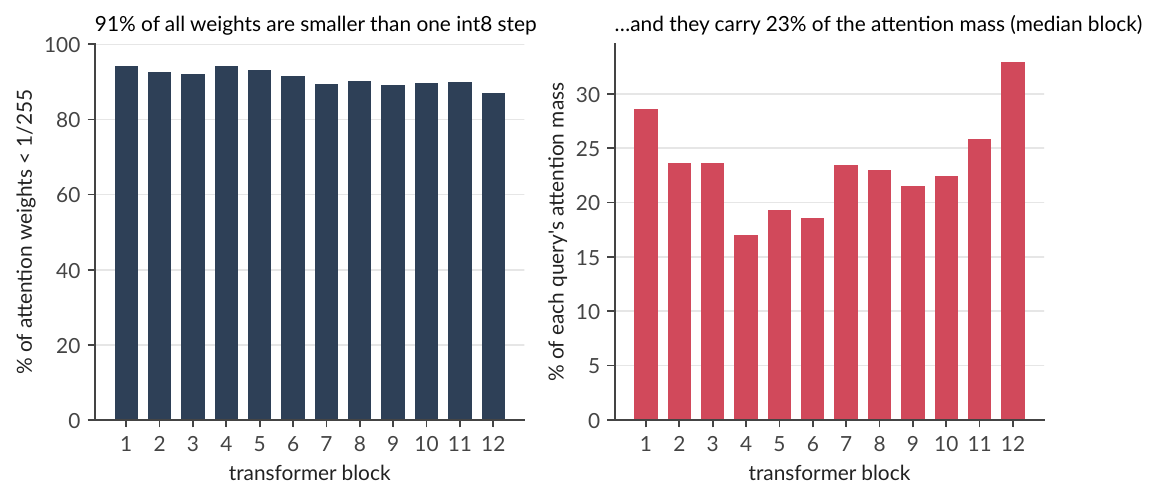}
\caption{Why attention breaks in int8, measured in float \dav-Small at the \hailo's 449-token input on 26 held-out frames.
Left: for each of the 12 transformer blocks, the percentage of softmax weights smaller than one int8 step ($1/255$); on
average 91\,\% are. Right: the share of each query's attention mass that these small weights carry (median 23\,\%), which
int8 rounds away. The last block, which feeds the depth head, loses the most.}
\label{fig:attn}
\end{figure*}

A softmax weight lies in $[0,1]$, and a uniform 8-bit quantiser divides that interval into steps of
$\Delta=1/255$. A head that spreads its attention evenly over all $N=449$ tokens gives each token a weight of $1/N$, which is
below one step:
\begin{equation}
\frac{1}{N}=\frac{1}{449}\approx2.2\times10^{-3}\;<\;\Delta=\frac{1}{255}\approx3.9\times10^{-3}.
\end{equation}
Such weights round to zero or to one step. Measured in float
\dav-Small at the chip's input on 26 held-out frames (\cref{fig:attn}), 91\,\% of all softmax weights are below one int8
step, and they carry 17--33\,\% (median 23\,\%) of each query's attention mass, most of it in the last block, which feeds
the depth head. The available remedies did not close the gap: re-calibration plus the compiler's quantisation fine-tuning
moved the share of frames with Pearson below 0.8 only from 48\,\% to 37\,\%, sensor-noise calibration models left
30--39\,\% broken, and Hailo's own Model Zoo build shows the same blockiness and object loss. Every int8 transformer build
we tested fails this way, and the attention-free student of \cref{sec:student} does not.

\section{An Attention-Free Student}\label{sec:student}
We distil \dav-Large into a network with no attention, no LayerNorm and no softmax: a ResNet-34 encoder~\cite{he2016resnet}
with ImageNet initialisation and a U-Net decoder~\cite{ronneberger2015unet} of convolution, batch normalisation, ReLU,
bilinear $2\times$ upsampling and skip concatenation. It has 24.5\,M parameters, almost exactly \dav-Small's 24.8\,M, so the
comparison isolates architecture from capacity. It maps 288$\times$512 RGB, the robot camera's aspect, to relative
disparity at full resolution.

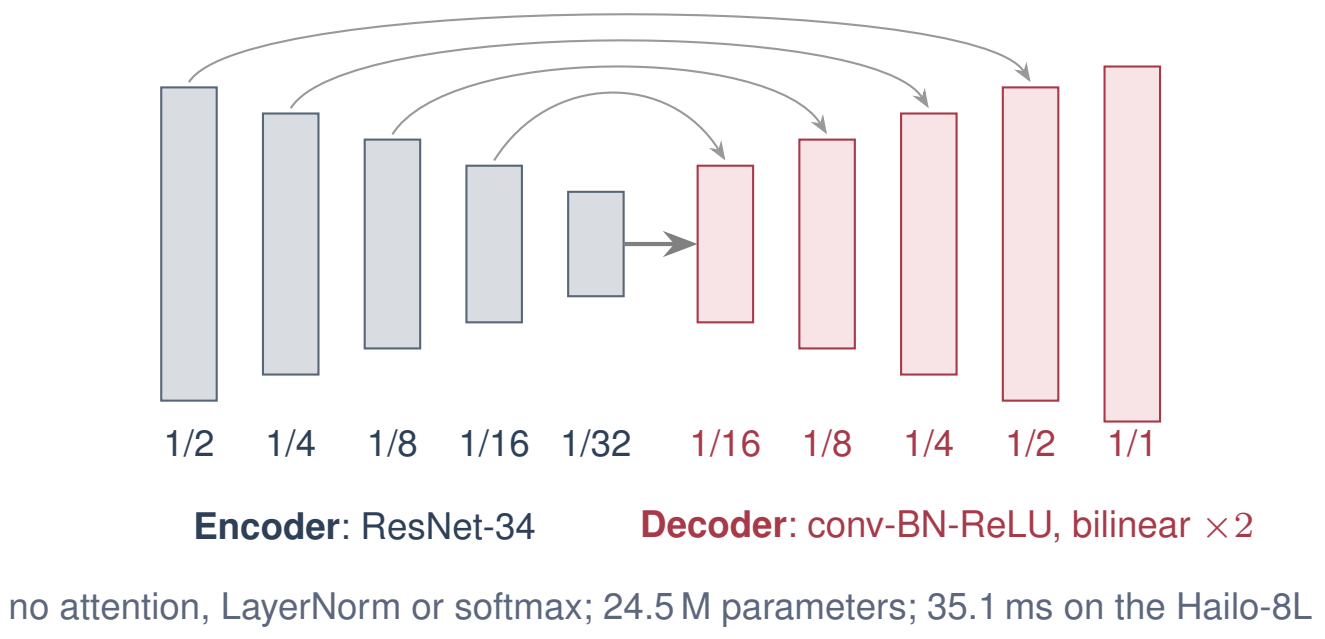
\begin{figure}[t]
\centering
\resizebox{\columnwidth}{!}{%
\begin{tikzpicture}[every node/.style={font=\sffamily\scriptsize}, x=0.62mm, y=0.7mm]
\foreach \x/\h/\l in {0/30/{1/2}, 11/25/{1/4}, 22/20/{1/8}, 33/15/{1/16}, 44/10/{1/32}} {
  \fill[chip!18] (\x,-\h/2) rectangle ++(6,\h); \draw[chip!80] (\x,-\h/2) rectangle ++(6,\h);
  \node[chip] at (\x+3,-19) {\l};}
\foreach \x/\h/\l in {58/15/{1/16}, 69/20/{1/8}, 80/25/{1/4}, 91/30/{1/2}, 102/34/{1/1}} {
  \fill[ours!15] (\x,-\h/2) rectangle ++(6,\h); \draw[ours!80!black] (\x,-\h/2) rectangle ++(6,\h);
  \node[ours!80!black] at (\x+3,-19) {\l};}
\foreach \a/\b/\h in {36/61/15, 25/72/20, 14/83/25, 3/94/30} {
  \draw[-{Stealth[length=1.3mm]}, gray!80] (\a,\h/2+0.5) .. controls (\a+6,\h/2+9) and (\b-6,\h/2+9) .. (\b,\h/2+0.5);}
\draw[-{Stealth}, thick, gray] (50,0) -- (58,0);
\node[chip, align=center] at (22,-27) {\textbf{Encoder}: ResNet-34};
\node[ours!80!black, align=center] at (85,-27) {\textbf{Decoder}: conv-BN-ReLU, bilinear $\times2$};
\node[mute, align=center] at (54,-35) {no attention, LayerNorm or softmax; 24.5\,M parameters; 35.1\,ms on the \hailo};
\end{tikzpicture}}
\caption{The attention-free student: a U-Net from 288$\times$512 RGB to disparity at full resolution (grey: skip connections).}
\label{fig:student}
\end{figure}

\keyline{Training.} The loss is the scale- and shift-invariant L1 of MiDaS plus four-scale gradient matching, so only the
shape of depth is supervised. v1 is distilled from teacher labels on 7{,}230 simulated point-of-view frames and 720 real
frames, with three scenes (crates, people, barrels) and the real robot video held out. v3 fine-tunes v1 for eight epochs on
\emph{rendered} ground-truth disparity for simulated frames (13{,}902 simulated + 2{,}024 real), which reduces the
teacher-inherited range bias on simulated props from about $-30$\,\% to $-7$\,\%. v5, the deployed student, weights the loss
four times higher on rows below the horizon, where obstacles meet the floor; it cuts spurious near-floor hits from 13.7\,\% to
10.2\,\%.

\keyline{On the chip.} The student compiles with normalisation folded into the HEF. Replayed on the real chip against its
own float weights on 500 frames, v3 reaches Pearson \chipfidpfifty{} at the median (5th percentile \chipfidpfive) and v5
\chipfidvfivepfifty{} (\chipfidvfivepfive). The compiler's emulator had reported a correlation of $-0.41$ for this graph, a false
alarm: an NPU build should be judged by replay on the chip. The student takes 35.1\,ms on the NPU and 37.0\,ms per frame end to
end, 27.1~FPS at batch~1 (59~FPS at batch~8), against 455\,ms on the Pi's four CPU cores.

\section{On-Chip Benchmark}\label{sec:bench}
\keyline{Protocol.} 400 simulated frames from 36 episodes of the three held-out scenes, with the renderer's metric depth,
and 100 held-out real robot frames without ground truth. Every \hailo{} model runs on the chip with its outputs read back.
Each model sees a centre crop at its own aspect; predictions are aligned to ground-truth disparity by per-frame least-squares
scale and shift~\cite{ranftl2022midas}. We report AbsRel and $\delta_1$ on all valid pixels and on \emph{obstacle} pixels:
pixels within 3\,m that, back-projected with the true pose, lie on a prop's collider. The obstacle region removes the
renderer's floor speckle and measures what the planner reads. On real video the metric is Pearson correlation to the teacher.

\begin{table}[t]
\centering\footnotesize
\setlength{\tabcolsep}{4pt}
\caption{Depth on held-out frames (400 sim frames with metric GT; $r$: median Pearson to \dav-Large on 100 real frames).
Bold: best on the chip.}
\label{tab:bench}
\input{tables/paper_bench.tex}
\vspace{1mm}\par\raggedright\footnotesize
AbsRel: mean absolute relative error (lower is better); $\delta_1$: share of pixels within 25\,\% of the true depth (higher
is better); all: every valid pixel; obst.: pixels on obstacles within 3\,m; real $r$: Pearson correlation with the teacher on
real video; Speed: frames per second on the Pi (chip rows) or milliseconds per frame on the GPU (float rows).
\end{table}

\begin{figure*}[t]
\centering
\includegraphics[width=\textwidth]{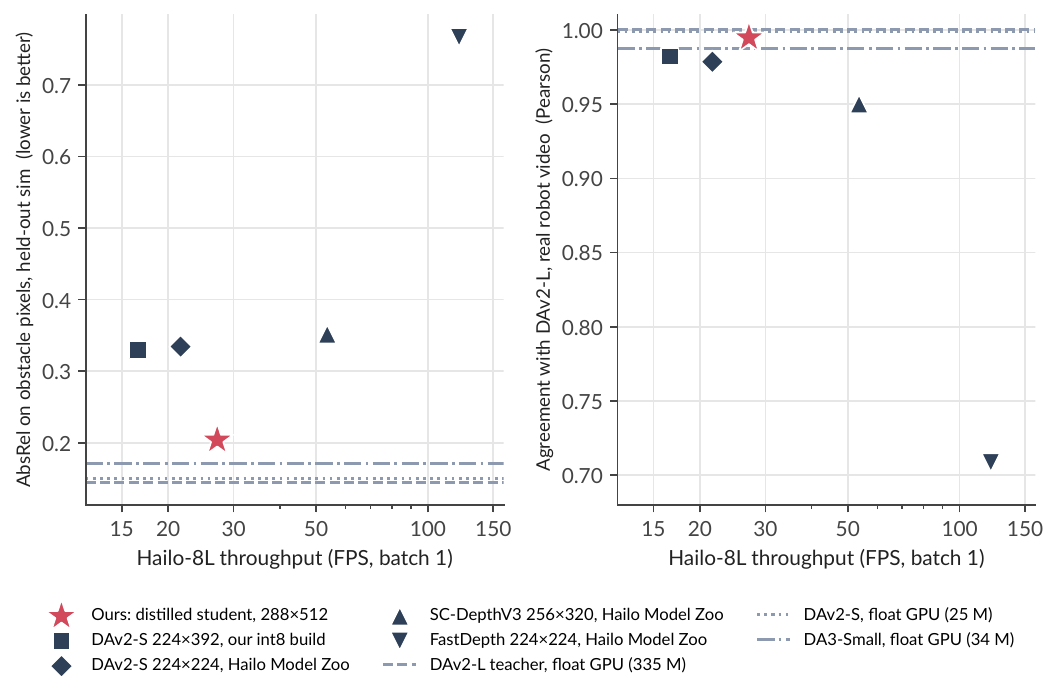}
\caption{Accuracy against measured on-chip throughput on the \rpi{} + \hailo{} (batch~1, logarithmic axis). Left: AbsRel on
obstacle pixels of the 400 held-out simulated frames (lower is better). Right: Pearson agreement with \dav-Large on 100 real
robot frames (higher is better). Markers: models running on the chip (red star: ours). Horizontal lines: float GPU models,
which cannot run on the Pi at these rates. The student is both the most accurate and faster than both \dav-Small builds.}
\label{fig:pareto}
\end{figure*}

\begin{figure*}[t]
\centering
\includegraphics[width=\textwidth]{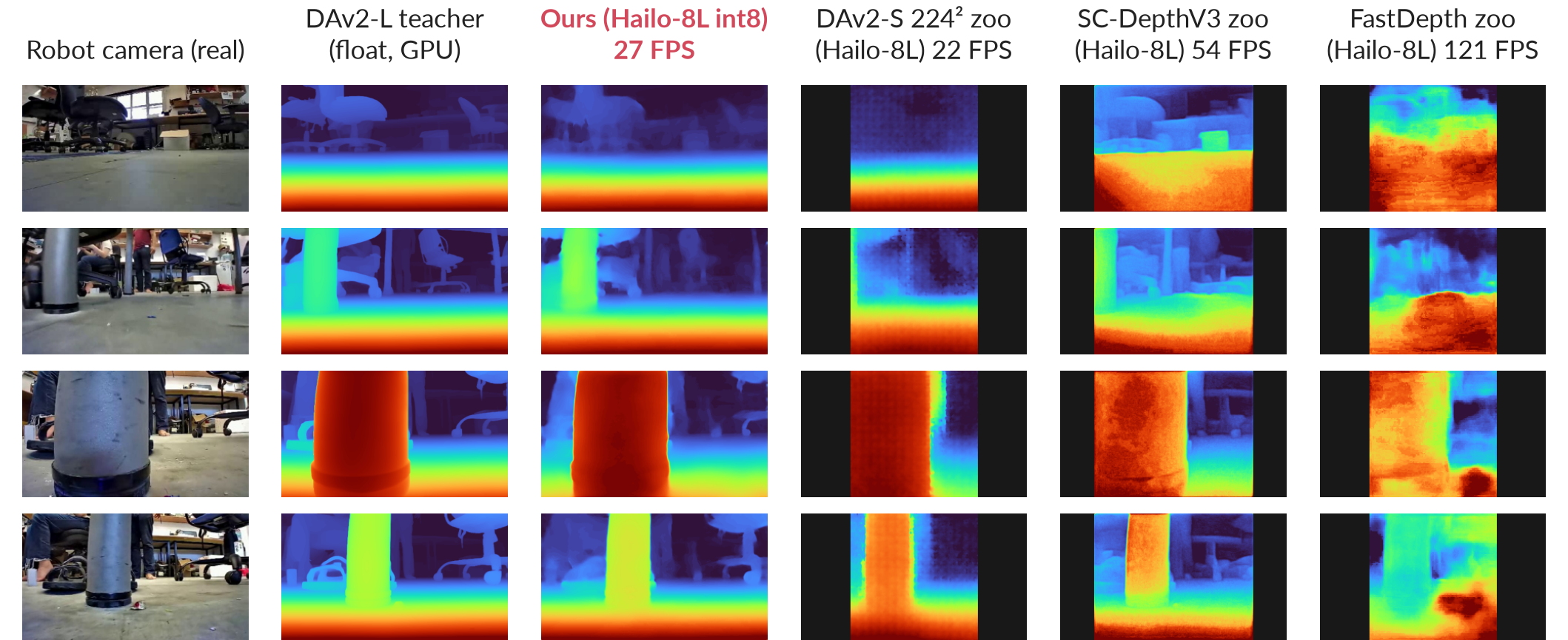}
\caption{Depth computed \emph{on the \hailo{}} for four real robot frames (rows). Columns, left to right: the robot's
camera image; the float \dav-Large teacher on a GPU (the reference); our attention-free student (int8, on the chip,
27~FPS); Hailo's Model Zoo build of \dav-Small; and the two self-supervised models Hailo ships, SC-DepthV3 and FastDepth.
Colour encodes relative depth (red near, blue far); dark bands lie outside a model's input crop. The student keeps the
teacher's structure; the int8 transformer is blocky and loses fine structure (the chairs in the first row); the small
zoo models are coarse.}
\label{fig:qual}
\end{figure*}

\keyline{Results} (\cref{tab:bench,fig:pareto}). On the chip the student is in a different class: overall AbsRel 0.269
against 0.354 for Hailo's \dav-Small, 0.414 for SC-DepthV3, 0.477 for our \dav-Small build and 0.530 for FastDepth, and
0.197 against 0.330--0.767 on obstacle pixels. On real video (\cref{fig:qual}) it tracks its teacher at $r=0.996$, where the vendor's \dav-Small
reaches 0.979 and FastDepth 0.709. Speed is not traded for this: at 27.1~FPS it is faster than both \dav-Small builds (21.6 and
16.6~FPS) while processing 2.9$\times$ and 1.7$\times$ more pixels. Against float foundation models the student gives up a
few points on obstacles (the teacher reaches 0.144) but is better overall, because it was fine-tuned on rendered ground truth of
this domain; Depth Anything~3 is not better than \dav{} here. The simulated frames come from one reconstructed yard; the real
frames show that the fine-tuning did not damage real-world behaviour, but agreement with a teacher is not accuracy.

\section{From Relative Depth to a Metric Scan}\label{sec:space}
The student's disparity is $d=a/Z+b$ with unknown, drifting $a$ and $b$. Instead of learning metric depth we take metre scale
from the camera height $h$. For a level camera with focal length $f_y$ and principal row $c_y$, a floor point at image row
$v>c_y$ lies at
\begin{equation}
Z_{\mathrm{floor}}(v)=\frac{f_y\,h}{v-c_y}.
\end{equation}
An obstacle meeting the floor at row $v_c$ is at $Z_{\mathrm{floor}}(v_c)$, whatever the network's scale. The network only
has to say where in each column the floor ends, a question about the shape of the depth map. The image columns become a
69$^\circ$, 4-metre ``laser scan''.

\begin{figure*}[t]
\centering
\includegraphics[width=0.85\textwidth]{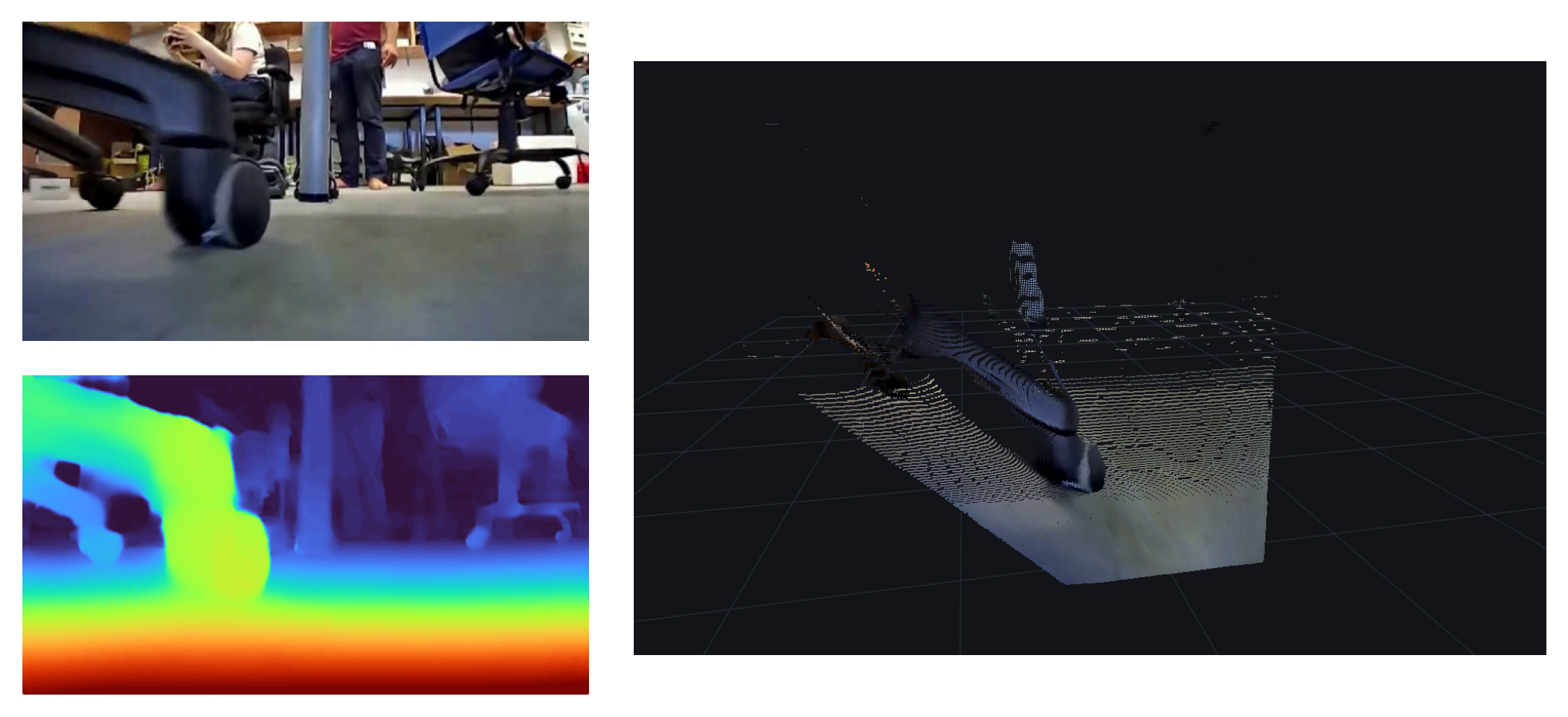}
\caption{One real camera frame lifted to metric 3D on the Pi. Top left: the robot camera image. Bottom left: relative depth
computed on the \hailo{} (int8, 35\,ms). Right: the resulting point cloud of about 80{,}000 points, seen from behind the robot
(grid: 0.5\,m). Metric scale comes from the floor profile and the camera height, with no metric training; pixels at depth
discontinuities are removed.}
\label{fig:cloud}
\end{figure*}

\begin{figure*}[t]
\centering
\includegraphics[width=0.85\textwidth]{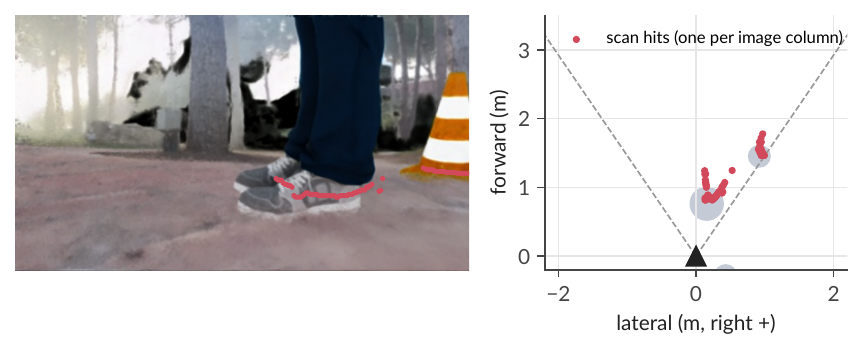}
\caption{The obstacle scan on a held-out simulated frame, computed from \hailo{} depth. Left: the robot's view with the
ground-contact point found in each image column (red), where a person's foot and a cone meet the floor. Right: the same hits
in bird's-eye view (metres, robot at the origin, dashed lines: field of view) against the true footprints of the props
(grey discs). The hits land on the footprints although the depth network has no metric scale.}
\label{fig:scan}
\end{figure*}

\keyline{The scan.} Per row, the 25th percentile of disparity over the central 80\,\% of columns models the floor $f(v)$,
made monotone from the horizon down. For a pixel, $(d-b)/(f(v)-b)=Z_{\mathrm{floor}}/Z$ is scale-free; a pixel at least
$H_{\min}=5$\,cm above the floor at its range exceeds $1/(1-H_{\min}/h)$, and the first run of five such rows in a column is
the obstacle's contact. A second, profile-free detector flags vertical faces, where the normalised disparity slope
$\partial_v d\,(v-c_y)/(d-b)$ drops from $\approx1$ (floor) to $\approx0$ (face). Following the contact upward gives an
obstacle's height; only hits at least 6\,cm tall are mapped, which cut phantom hits from 2{,}376 to 75 on six recorded
episodes. Two fixes came from failure analysis: anchoring the floor profile at the horizon (a chair seat above the camera had
become ``floor'', making a tube at 0.68\,m read 1.16\,m), and a detector for floating overhangs such as seats and table tops
(\cref{fig:scan}). The same profile is a ruler for every pixel, $Z=Z_{\mathrm{floor}}(v)(f(v)-b)/(d-b)$, which gives a metric
point cloud from one frame without metric training (\cref{fig:cloud}); pixels at depth discontinuities are dropped. Per pixel it
is less robust than the contact scan (median error $-2$ to $+8$\,\% on obstacle pixels, but a long tail at object edges), so we
use it for overhangs and visualisation, not for ranging.

\keyline{Map and limit.} Hits accumulate in a 5\,cm world grid; a cell becomes an obstacle after three hits and if it holds at
least 30\,\% of the hits of the densest cell within 0.2\,m. The remaining perception error is low structure: at the deployed
5\,cm threshold the scan sees prop sides 5.5\,cm (ground-truth depth) and 5.1\,cm (student) inside the true footprint. Lowering
the threshold to 2\,cm closes the gap for ground-truth depth ($-0.5$\,cm) but not for the student (+2.5\,cm), whose
floor-to-base boundary is blurred.

\section{Onboard Autonomy}\label{sec:autonomy}
\keyline{Planning.} A hybrid A$^\ast$~\cite{dolgov2010hybrid} over rear-axle poses (0.1\,m cells, 72 headings, ten 0.25\,m
primitives forward and backward) checks a three-circle footprint against the sensed map and a prior drivability map of the
yard. Its heuristic is the larger of an obstacle-aware 2D geodesic distance and a precomputed table of the obstacle-free
non-holonomic cost of the lattice; without the table, a goal behind the car flooded 40{,}000 expansions. Primitives are planned
at 75\,\% of steering lock (1.17\,m radius), because the car turns wider than its bicycle model and plans at full lock
saturated the follower. A cascade relaxes the clearance margin from 0.10 to 0.06\,m when a search fails and widens it to
0.20\,m when that costs at most 1.6$\times$ the length. Cells the car already overlaps are passable only within 0.3\,m of the
start, so it can back out of a contact without planning through a touched crate face. Searches run in a worker process:
110\,ms at the median and 547\,ms at the 90th percentile on the Pi.

\keyline{Control and safety.} Pure pursuit tracks one direction segment at a time. The car re-plans when the path is
blocked, when it leaves the path, or when it is wedged (throttle without progress for 1.5\,s, which also places a ``bump''
obstacle at the bumper). Every reverse passes the Traxxas ESC's brake-then-neutral gate, which the simulator models.

\begin{figure*}[t]
\centering
\includegraphics[width=0.9\textwidth]{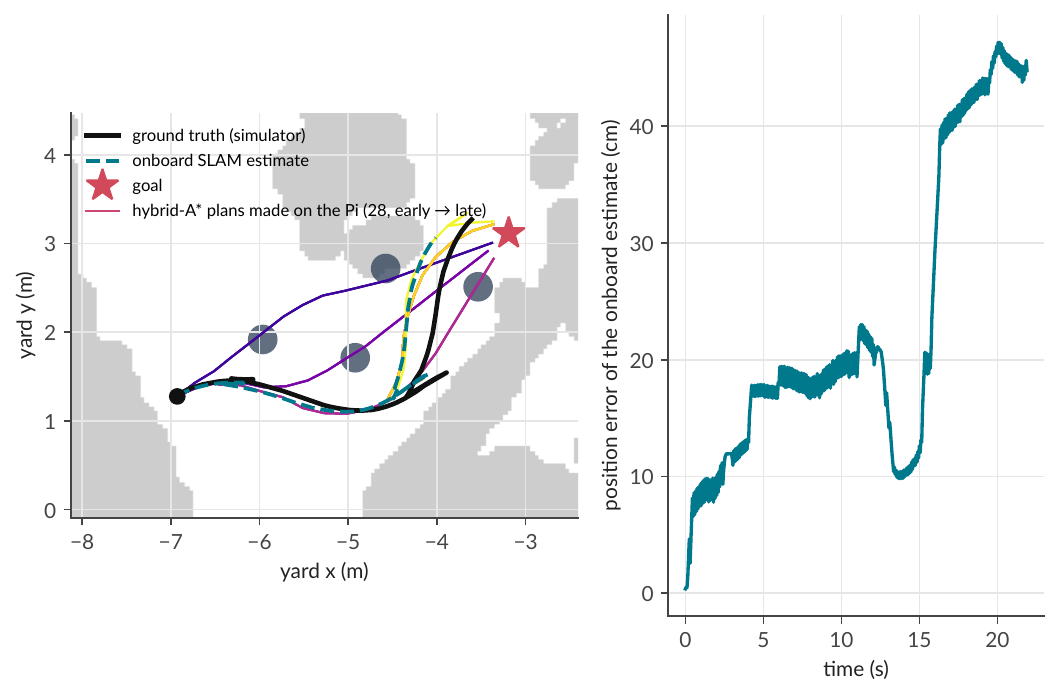}
\caption{Onboard SLAM and global planning on the robot's own pose, in one cone-slalom episode. Left: top view of the yard
(metres), with cones (dark discs), ground the prior map marks as non-drivable (grey), the true path (black), the onboard SLAM
estimate (dashed), the goal (star) and every hybrid-A$^\ast$ plan the Pi made (thin lines, early to late). Right: the
distance between the SLAM estimate and the true position over time. The car reached the goal despite 0.47\,m peak drift.}
\label{fig:slam}
\end{figure*}

\keyline{SLAM.} About 400 floor features within 1.5\,m are tracked with forward--backward Lucas--Kanade~\cite{lucas1981lk},
lifted to metric floor points with the same camera-height geometry, and fit by a RANSAC 2D rigid transform. Wheel odometry
(encoder and gyro) enters directly. A slip guard declares the car still when the rotation-compensated floor flow is below
2\,px while the encoder claims at least 2\,cm per frame. Offline on 50 recorded episodes (\cref{tab:odo}) an IMU alone barely helps; the wheel encoder is what makes
own-pose navigation work. In the closed loop, with encoder and gyro, the onboard estimate stays within 0.13\,m of the truth at
the median and 0.99\,m at the 90th percentile over ten seeds (\cref{fig:slam}).

\begin{table}[t]
\centering\footnotesize
\caption{Self-localisation alternatives, offline on 50 recorded 35-s episodes (absolute trajectory error, m).}\label{tab:odo}
\begin{tabular}{@{}lrr@{}}
\toprule
Odometry source & median & p90 \\
\midrule
Ground-plane VO (camera only) & 0.42 & 1.32 \\
VO translation + gyro heading & 0.36 & 1.04 \\
Encoder + VO heading & 0.22 & 1.18 \\
Encoder + Ackermann heading from steering & 0.25 & 0.61 \\
\textbf{Encoder + gyro} & \textbf{0.08} & \textbf{0.19} \\
\bottomrule
\end{tabular}
\vspace{1mm}\par\raggedright\footnotesize
Median and 90th-percentile absolute trajectory error over the 50 episodes, in metres (lower is better). VO: visual
odometry from the floor; encoder: wheel encoder; gyro: yaw rate.
\end{table}

\begin{figure*}[t]
\centering
\includegraphics[width=0.85\textwidth]{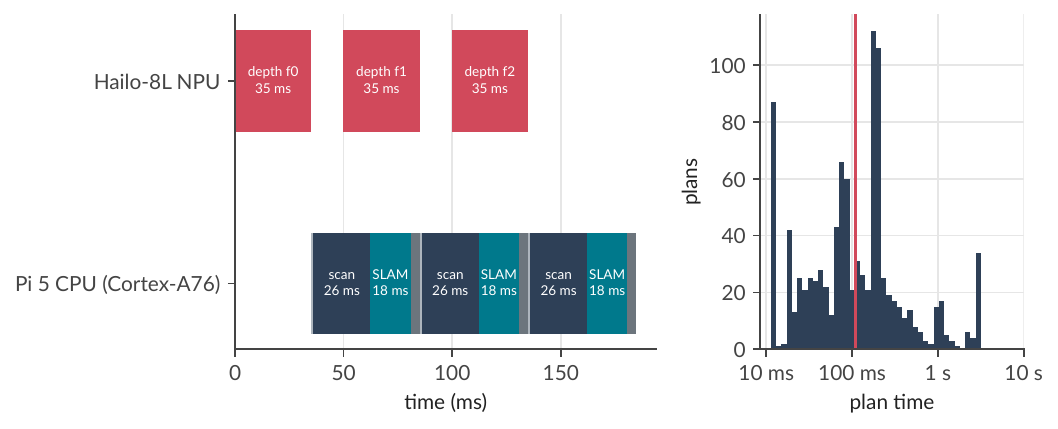}
\caption{The real-time budget on the \rpi{} + \hailo{}, measured on the robot computer. Left: median duration of each
per-frame stage over three consecutive frames; the NPU computes depth for frame $k+1$ (top lane) while the CPU runs the
decoding, scan and map, SLAM and navigation for frame $k$ (bottom lane). Right: histogram of hybrid-A$^\ast$ search times on the Pi
(logarithmic axis; red line: median, 110\,ms). Planning runs in a separate process and does not block the frame loop.}
\label{fig:timing}
\end{figure*}

\keyline{Real-time budget.} With the deployed code on the real board the NPU takes 35.0\,ms per frame and the CPU stages
(scan, map, SLAM, navigation) 46.7\,ms at the median (64.1\,ms at p90), overlapped (\cref{fig:timing}). On real 1280$\times$720 video the
standalone runner reaches 16.5\,Hz with SLAM and 27\,Hz without it, limited by the NPU. Over 1000 episodes and 28\,h the NPU time
did not drift (p99 35.9\,ms) and the firmware's throttle register stayed clear. The \hailo{} is rated at about
1.5\,W~\cite{rpi2024aikit}; the module exposes no power sensor, so we report no measured power.

\section{Hardware-in-the-Loop Evaluation}\label{sec:hitl}
\keyline{Set-up.} Isaac Sim with Isaac Lab~\cite{nvidia2025isaaclab} renders a Gaussian-splat reconstruction~\cite{kerbl2023gs,
wu2025_3dgut} of a real yard with a heightfield collider. A model of the project's Traxxas drives with $\pm0.349$\,rad steering
and up to 0.72\,m/s; its level camera is 504$\times$280 with a 69.4$^\circ$ field of view, 0.13\,m above the ground. Eleven
prop types (chairs, a cone, a box, a barrel, a bin, a sign, a crate, two people) carry colliders sized to what the car can
hit. Ten frozen scenes put 3--6\,m between start and goal; three were never seen by the depth network. Every frame is sent to
the robot computer, and each returned command is applied only when it arrives. An episode is \emph{clean} if the goal is reached
(within 0.45\,m, in 35\,s) and no point of the car's real 0.557$\times$0.294\,m footprint ever touched a collider. Every arm runs
at least five seeds, each jittering the start.

\keyline{The board in the loop.} The first arms ran on the real Pi and \hailo{}. To run arms in parallel, the identical Pi code
also runs on the laptop with replies padded to the Pi's measured round trip and plans slowed to the Pi's speed. The deployed
configuration was then re-run on the real board: 8.2~$\pm$~0.4 clean against 7.6 for the laptop twin on the same seeds, with a
median round trip of 85\,ms (p99 113\,ms) over 9{,}371 replies against the twin's 84\,ms.

\begin{table}[t]
\centering\footnotesize
\caption{Clean goals of 10 (mean $\pm$ sd over $n$ seeds) and goals reached.}
\label{tab:hitl}
\input{tables/paper_hitl.tex}
\vspace{1mm}\par\raggedright\footnotesize
Clean: goals reached with no contact between the car's footprint and any obstacle, out of 10 scenes. Reached: goals
reached with or without contact. $n$: number of seeds. GT pose: the simulator supplies the car's position; own pose: the
car estimates it with its onboard SLAM. Bold: the deployed configuration.
\end{table}

\begin{figure*}[t]
\centering
\includegraphics[width=0.495\textwidth]{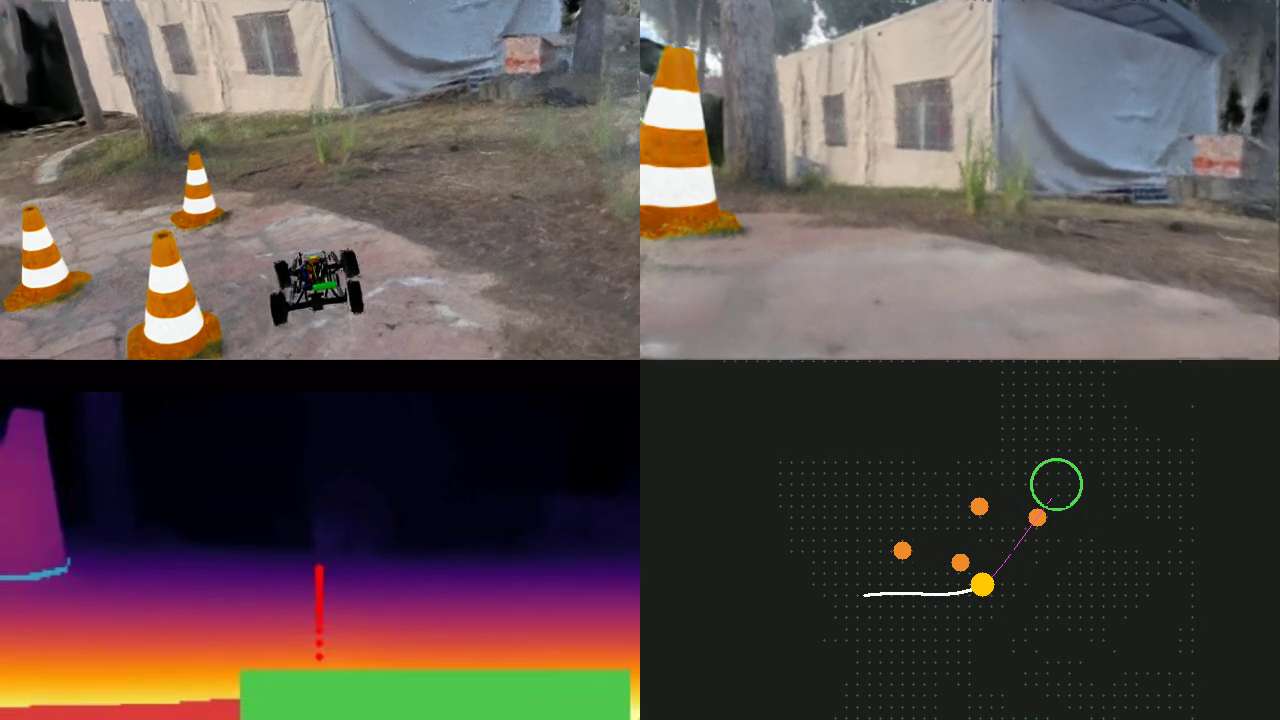}\hfill
\includegraphics[width=0.495\textwidth]{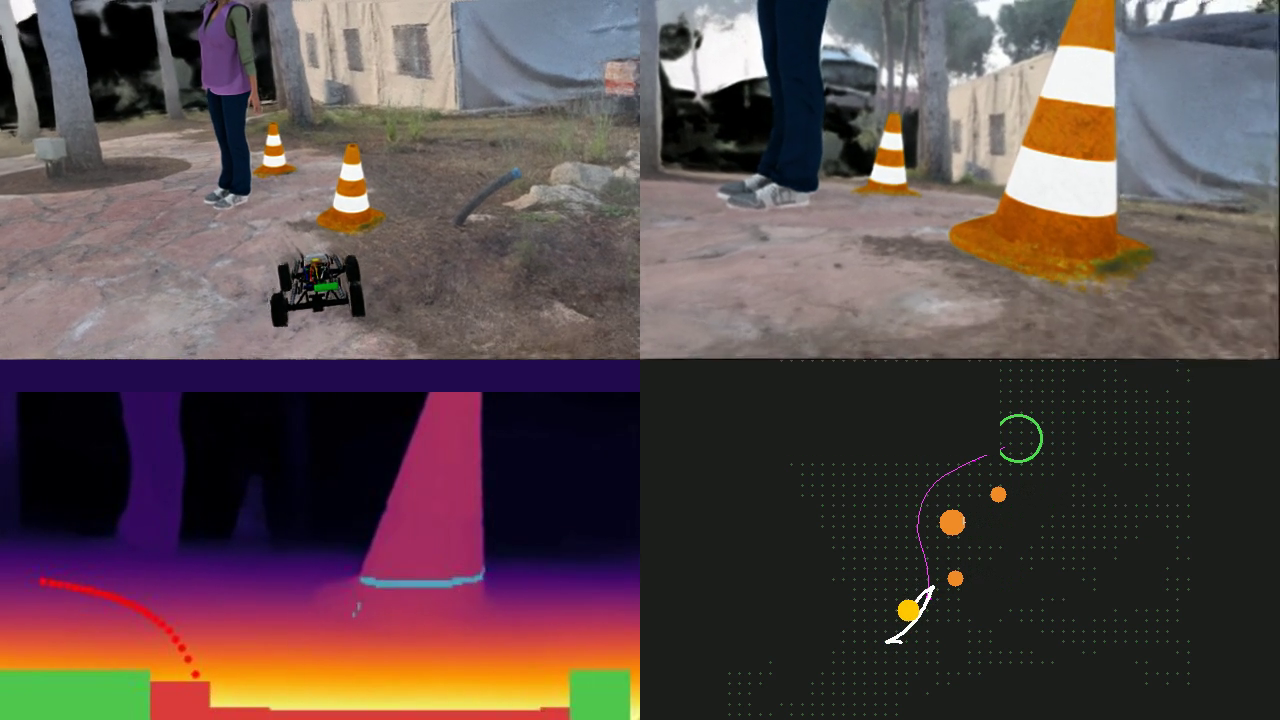}
\caption{The deployed stack driving with the real Pi + \hailo{} in the loop, on its own SLAM pose. Left: cone slalom. Right:
backing away from a cone seen at close range. In each image: top left, a chase camera; top right, the robot's camera;
bottom left, depth computed on the chip with the commanded steering arc (red); bottom right, a bird's-eye map with the
obstacles (orange), the goal (green circle), its SLAM path (white), its current plan (magenta) and the car
(yellow).}
\label{fig:hitlshots}
\end{figure*}

\begin{figure*}[t]
\centering
\includegraphics[width=0.9\textwidth]{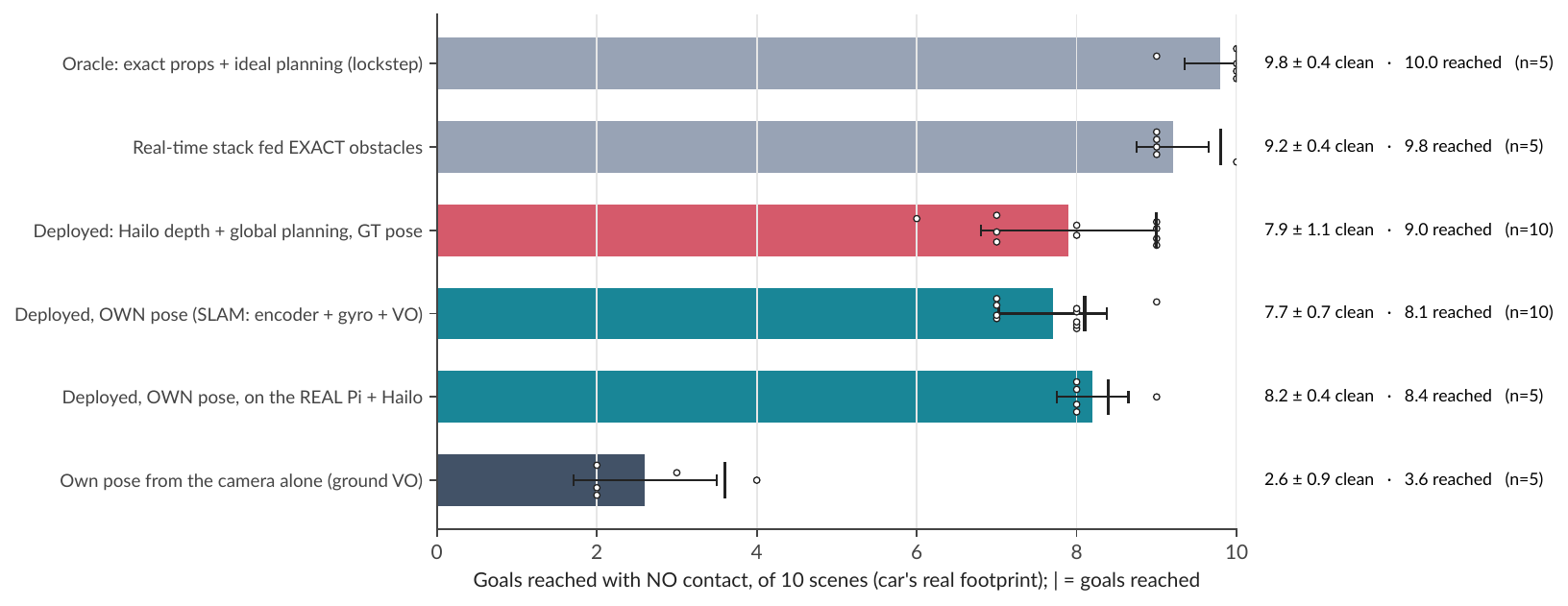}
\caption{The error budget: goals reached with no contact, out of 10 scenes, from an oracle down to camera-only odometry. Bars:
mean over seeds; dots: individual seeds; error bars: one standard deviation; vertical ticks: mean goals reached with or
without contact. The drops between the first four bars give the cost of each component: execution (0.6), perception (1.3) and self-localisation (0.2 clean, within seed noise; 0.9 reached).}
\label{fig:hitl}
\end{figure*}

\keyline{The error budget} (\cref{tab:hitl,fig:hitl}). An oracle that knows every prop and plans in lockstep reaches 9.8:
a perfect score is feasible. The same real-time stack fed exact obstacles reaches 9.2, so latency, the follower and the ESC gate
cost about 0.6. With depth from the \hailo, the deployed stack reaches 7.9~$\pm$~1.1 (9.0 reached): perception costs about 1.3.
On its own pose with simulated encoder and gyro it reaches 7.7~$\pm$~0.7; self-localisation costs 0.2 clean goals, within seed noise, but 0.9
goals reached. Camera-only odometry reaches 2.6, so a wheel encoder is not optional for this car. With simulator pose most misses
reach the goal after a graze (11 of 21); on own pose most are timeouts (18 of 23).

\keyline{v3 or v5.} The floor-weighted v5 replaced v3 on modest evidence, which we state exactly: with simulator pose the two
are not resolved (7.9 vs 7.9 at ten seeds), with per-scene differences in both directions; on own pose v5 leads, 7.7 against 7.1, not
significantly (permutation test $p\approx0.17$). It costs nothing on the chip and has the lower obstacle AbsRel.

\begin{figure*}[t]
\centering
\includegraphics[width=0.9\textwidth]{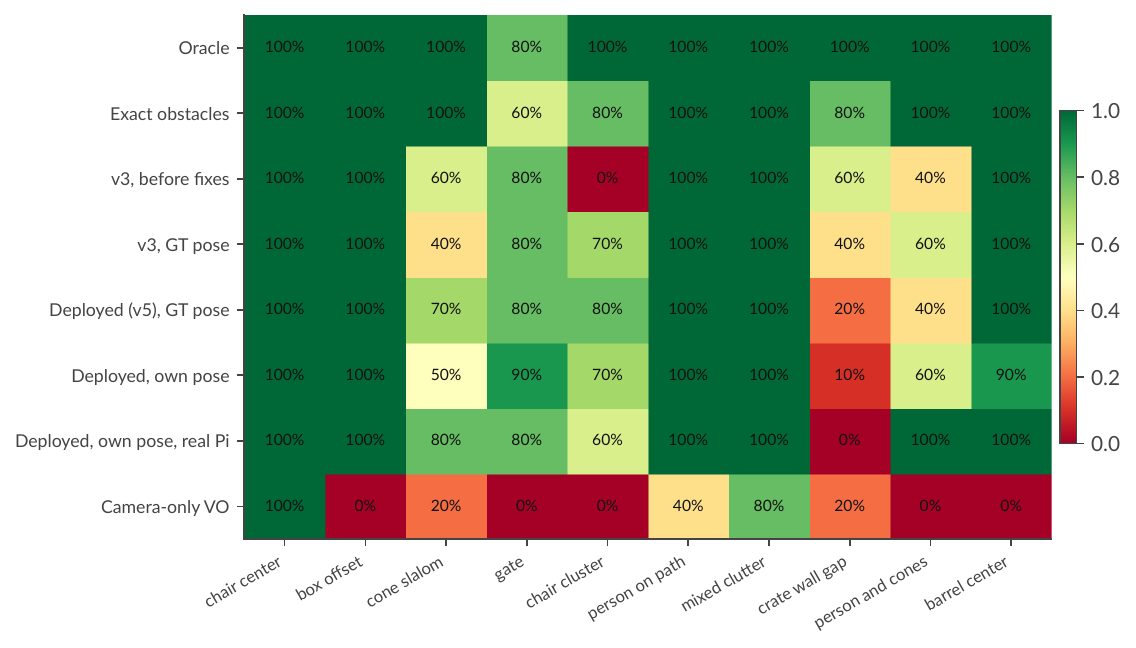}
\caption{Clean success per scene: for each configuration (rows) and each of the ten scenes (columns), the percentage of
seeds that reached the goal with no contact (green high, red low). Failures concentrate on the cone slalom, the gate, the
chair cluster, the crate wall and the person-and-cones scene; camera-only odometry fails in most scenes.}
\label{fig:scenes}
\end{figure*}

\keyline{What remains, and what did not help.} The remaining contacts are almost all the car's front corner grazing the side of
a cone, a person or a crate that the scan places 5--15\,cm too far out: the low-base limit of \cref{sec:space}. In one scene the
goal starts 63$^\circ$ off axis and a crate lies outside the $\pm34^\circ$ view, so first plans run through unseen space; this is
an exploration problem. More than a dozen interventions moved contacts within the seed noise and are reported as such
(\cref{tab:negative}).

\begin{table}[t]
\centering\footnotesize
\setlength{\tabcolsep}{3pt}
\caption{Interventions that did not improve clean success ($\geq$5 seeds; against the baseline of the time on the same seeds).}
\label{tab:negative}
\begin{tabular}{@{}p{3.75cm}lr@{}}
\toprule
Intervention & Outcome & Clean \\
\midrule
+6\,cm skirt on small obstacles & slalom closes & 7.2 vs 7.6 \\
Orange-mask cone contacts & slalom timeouts & 7.8 vs 7.6 \\
Look-first arc, off-axis goal & neutral, own pose & 6.8 vs 7.1 \\
Third planning pass & contacts move & 7.8 vs 8.0 \\
Faster follower & drift p90 $\times1.6$ & 7.4 vs 7.6 \\
Uniform blocked-path margin & re-plans stall & 6.6 vs 7.4 \\
2$\times$ camera pixels & no change & 2.6 vs 2.6 \\
Renderer depth for the student & floaters, speckle & 5.2 vs 6.8 \\
\bottomrule
\end{tabular}
\vspace{1mm}\par\raggedright\footnotesize
Clean: goals reached with no contact, out of 10, with the intervention vs without it on the same seeds. None improved
clean success.
\end{table} One comparison is worth stating plainly: the student drives better than the
simulator's own depth buffer (6.8 vs 5.2), whose splat floaters and floor speckle corrupt ground-contact ranging, even though
the renderer's arm ran without latency.

\section{A Real Drive: Depth and SLAM against the RealSense}\label{sec:real}
The car was driven by hand through a sunlit park while it recorded every D435 stream (colour, on-chip stereo depth, both
infrared images) and the student's \hailo{} output for every frame: 219\,s recorded (SLAM window
216\,s). Three recorder defects (a header byte shift, stereo timestamps drifting $\approx$8\,\%,
a power cut at the end) were worked around before any comparison (49 of 388 evaluation frames excluded).

\textbf{Which depth is better.} The drive has no lidar, so the reference is agreement between three different matchers on the
same infrared pair (D435 ASIC, SGBM, Fast-FoundationStereo; within $2\,\%+1\,\%/\mathrm{m}$, 0.3--3\,m). Fast-FoundationStereo's
backbone includes \dav{} features, but as one voter it can only veto points. Points are registered into the colour view with the D435's factory calibration, read after the drive: the
nominal values misplaced every point by a median 14.5\,px; on the same points the factory values cut 17 of 19 networks' AbsRel by
10--47\,\% (stereo unchanged) and align depth with colour edges better on all 20 scorable frames.
1{,}000 points are frozen once and every monocular method is scored on them, relative models aligned per frame on
held-out points (Table~\ref{tab:real}); CIs are frame-cluster bootstraps, re-checked with 10\,s blocks as units. The student,
live on the chip, reaches AbsRel 0.262 and is not resolvably different from the Hailo model-zoo \dav-S (paired difference
$+0.017$, 95\,\% CI $[-0.060, +0.118]$, $p=0.70$; $|\log(\hat z/z)|$ $+0.009$, CI $[-0.012, +0.032]$): on these
mostly-ground points the interval bounds any advantage of ours at 0.060 (0.040 with blocks), against the simulator's 0.084 (CI 0.060--0.113 over
104 runs; a different metric), so the block bound lies below that interval; in relative terms, $-8$\,\% (CI $-30$ to $+23$\,\%)
against the simulator's 18--30\,\%, one drive cannot decide; on the two-matcher set below neither bound holds. On an obstacle stratum (points
$\ge$0.15\,m above the floor plane; 319 points, 87 frames; added after the main null), $|\log(\hat z/z)|$ leans the
simulator's way, ours against the zoo model $-0.165$, but with blocks exact $p=0.046$ (Holm 0.046--0.14 by family; exploratory) and $-0.123$, $p=0.055$, on the
two-matcher set; the median barely differs (0.232 vs 0.240; 0.369 vs 0.346 two-matcher), and with one scale per drive ours is worse there ($+0.334$,
$p=0.040$, Holm 0.058). \dav-L (not on the two-matcher set) and DA3-L are better on obstacles by 0.153 and 0.184. Without Fast-FoundationStereo (a new 1{,}000-point set on 321 frames) the mean leans our
way ($-0.061$, block CI $[-0.130, +0.007]$, $p=0.078$), carried by the zoo model's blow-ups ($|\log|$: $-0.006$). Live, the student is worse than its own chip replay by $+0.004$ (CI $[0.001, 0.006]$; not on the two-matcher set). It
beats the other three edge builds (block $p\le0.014$; SC-DepthV3 and FastDepth get a median scale only) and trails its teacher
\dav-L by 0.085 per frame (CI $[0.047, 0.141]$). GPU metric models score 0.20--0.22 aligned (not resolvably better
than ours) and overestimate distance from 16\,cm by $\times1.7$--$3.5$ raw. With one scale per drive MoGe-3/2 lead (0.26) and the relative networks span 0.36--0.44 (ours 0.410): their scale moves
from frame to frame, so the robot takes a per-frame scale from camera height (unscored).

\textbf{Same poses, swap the depth.} As a pseudo-lidar at the same poses, where the onboard rule flags an obstacle base beyond
1\,m the two depths differ by a median 25\,\% (931 frames; 26\,\% with the factory principal point), but 85\,\%
of the stereo's flags sit within 5\,cm of its fitted floor (the grass surface): the gap compares where the rule fires. Free of
the rule, the network's floor-fit metric depth at the measured horizon reads raised structure nearer beyond 1.5\,m (CIs 0.66--0.92,
0.57--0.79), unresolved at 0.6--1.5\,m, weakly farther at 0.3--0.6\,m; depth and scale are not separable here. Replayed open loop with the measured mount, the planner
stops on 37\,\% (block CI 23--50\,\%) of frames with the network's depth and 83\,\% (75--90\,\%) with the stereo's; ground flags are implicated in at most 84--87\,\% of those stops. No stricter rule variant (11 tried) keeps the simulator's props in view: each loses at least 22 percentage points of recall. A stricter vertical-face test cuts the held-out half's stops from 47\,\% to 17\,\% but keeps only 78\,\% of the raised flags (bar: 80\,\%), so it fails too.

\textbf{SLAM.} The reference is RTAB-Map on the D435 infrared image and stereo depth: one graph over 97\,\% of the drive, four
loop closures, 47 odometry resets. Skill ($1-\mathrm{RMS}/\mathrm{RMS}$ of standing still) measures agreement with this
reference, not accuracy: one drive; one run per method except ORB-SLAM3 stereo, DROID and VGGT (three runs each). The
reference's re-runs differ from it by 0.64--0.79\,m RMS and from each other by 0.33--0.49\,m; against any of four reference runs no skill moves by more than 0.06 and no
full-coverage ranking changes. Sim(2) skill cannot fall below 0, SE(2) skill can;
under Sim(2) ours scores $+0.10$ and the commands $+0.08$: a 0.02 gap from one run, not evidence of a lead. VGGT-SLAM~2.0
reaches $+0.80$ (three bit-identical runs; 117\,s in keyframe gaps, $+0.79$ at keyframes only); DROID-SLAM $+0.27\pm0.19$ (three runs; it does not rank
against DPV-SLAM's $+0.39$); CoreSLAM on a RealSense scan $+0.02$ rigid ($+0.06$ Sim(2)); the commands alone, without
vision, $-0.31$. ORB-SLAM3 stereo reaches $+0.879\pm0.013$ over three runs on the 13--19\,\% it tracks, but skill on a partial overlap does not rank against
skill over the whole drive, and it shares the reference's infrared images. Our onboard SLAM scores $-0.11$ as deployed: its
ground odometry's speed gate was set from the \emph{simulated} car and rejected the fast 23\,\% of frames, so its path is
0.43$\times$ the reference's. With a physical gate the path is 1.08$\times$ (the only check of the reference's scale that
shares none of its inputs) but the skill falls to $-0.38$ to $-0.62$, worse than the commands alone (also under Sim(2)). The uncorrected mount (3.8$^\circ$ down,
roll $-1.4^\circ$) is the prime suspect, not isolated. As deployed, depth barely enters the trajectory:
with either gate, switching the scan match off changes neither arm by more than 0.015, so their gap is the odometry on different frame sets (5{,}612 vs 5{,}882), and this drive cannot rank the depths.

\begin{table*}[t]
\centering\footnotesize
\setlength{\tabcolsep}{6pt}
\caption{Real drive, 1{,}000 stereo-verified points (park, sun; one run; D435 factory calibration). AbsRel with the scale fitted per frame and once per
drive; $\Delta$ = ours (live) $-$ method in per-frame AbsRel, frame-cluster bootstrap 95\,\% CI. $^\dagger$Each matcher against the
other two on its own points: agreement, not comparable with the monocular rows. $^\ddagger$Median scale only (the others: scale and shift).}
\label{tab:real}
\begin{tabular}{@{}lcccc@{}}
\toprule
Method & AbsRel & per drive & $\delta_1$ & $\Delta$ [95\,\% CI] \\
\midrule
D435 / SGBM / FFS stereo$^\dagger$ & 0.009--0.040 & -- & 0.97--1.00 & -- \\
DA3 S--L (GPU) & 0.101--0.115 & 0.390--0.418 & 0.86--0.90 & $+0.129$ to $+0.143$ \\
DAv2 S/B/L (GPU) & 0.160--0.181 & 0.422--0.427 & 0.82--0.84 & $+0.076$ to $+0.092$ \\
Hailo zoo DAv2-S (chip) & 0.234 & 0.432 & 0.74 & $+0.017$ [$-0.060$, $+0.118$] \\
Ours, chip replay & 0.258 & 0.413 & 0.74 & $+0.004$ [$+0.001$, $+0.006$] \\
Ours, live on the chip & 0.262 & 0.410 & 0.73 & -- \\
SC-DepthV3 (chip)$^\ddagger$ & 0.392 & 0.435 & 0.22 & $-0.159$ [$-0.242$, $-0.038$] \\
FastDepth (chip)$^\ddagger$ & 0.394 & 0.415 & 0.23 & $-0.160$ [$-0.243$, $-0.041$] \\
DAv2-S, our int8 (chip) & 0.443 & 0.439 & 0.58 & $-0.178$ [$-0.229$, $-0.128$] \\
\bottomrule
\end{tabular}
\end{table*}

\textbf{Which depth is better inside SLAM?} The real drive cannot say; the HITL simulator can, with ground truth. We added a
D435-geometry infrared pair (SGBM), a planar lidar ring and exact depth, drove one loop twice under three lighting presets and ran
every source through our SLAM (ATE after a rigid fit, mean $\pm$ sd over the three sessions). One trajectory:
the spread is lighting and render noise. The student is in-distribution here: it
was fine-tuned on exact disparity rendered in this same yard. The simulated car executes its commands without slip, so the
commands alone reach 0.107\,m (noon). Every arm therefore measures how much a depth source corrupts
a near-exact prior, and every camera arm except true-scale exact depth is worse than no vision. The deployed front end ignores depth (its tight visual-odometry prior: 0.60\,m whatever the source). With scan
matching as the odometry, the deployed floor-contact pseudo-scan from \emph{exact} depth reaches only $0.84\pm0.03$\,m (ours
$0.65\pm0.33$). Read off the depth image's horizontal plane, exact depth reaches $0.50\pm0.07$\,m with the per-frame floor-fit
scale. Smoothing that scale does not recover the loss ($0.42\pm0.13$), while the true metric
scale cuts the error 6.8$\times$ (6.3--7.4$\times$ per session), to $\mathbf{0.074\pm0.003}$\,m; on seven of ten further start seeds (three excluded by pre-stated checks), 6.5$\times$ (CI 5.5--7.7), and 6.3$\times$ (5.7--6.9) on the same loop driven the other way. Under the same floor-fit scale our network's error over exact depth's is
0.75$\times$ at noon (CI 0.49--1.14, at most 14\,\% worse) and, on those drives re-rendered, 0.66$\times$ at evening (0.57--0.78, lower) and 0.97$\times$ at dusk (0.70--1.36, at most 36\,\% worse). Even with a per-frame affine scale fitted to exact depth, at noon it stays 3.2$\times$ exact depth's true-scale error (CI 2.4--4.2), and 5.3$\times$ (4.1--6.7) fitted on the scan's rows: none of the three affine scales tried (floor fit, image-wide or scan-row fit) closes the gap. On exact depth the floor-fit scale error at 1--2\,m drifts slowly (sd 0.32 across frames, 0.04 frame to frame), so
smoothing cannot remove it. The lidar ring ($0.083\pm0.013$\,m) samples the same noise-free depth: a consistency check, not a lidar result. The floor fit suits avoidance, not mapping. Of the possible scale sources only stereo was
tested (true-scale SGBM $0.195\pm0.048$\,m; an IMU and a metric model untested). The
same network's int8 output from the real \hailo{}, replayed offline over the session frames, agrees with the float twin at a median
$r=0.9999$ per frame (dusk 0.9988; 2{,}652 frames), but its floor-fit scale differs by $-8$ to $+5$\,\% (dusk, 5--95th pct), which plausibly moves per-session ATE by a median 0.13\,m (up to 0.48\,m, 12 pairs): this SLAM amplifies tiny depth
differences, so only large effects are resolved, and smoothing's effect on the network flips sign between the two builds.

\section{Discussion}
\keyline{Distillation, not better quantisation.} The common assumption is that a foundation model reaches a small NPU through
better quantisation. On the \hailo{} it does not: int8 attention is structurally lossy at a few hundred tokens, for our build
and the vendor's. Distilling into a purely convolutional student turns a quantisation problem into a training problem, and
quantisation cost then disappears from the error budget (int8 = float at $r=$~\chipfidvfivepfifty). The plain recipe
(ResNet-34, U-Net decoder, scale-invariant loss, rendered-truth fine-tune) transfers to any integer-only accelerator.

\keyline{Metric navigation without metric depth.} Ground-contact ranging asks the network only where each column's floor
ends; the camera height does the rest. It is immune to relative depth's frame-to-frame scale drift, and its failure modes are geometric and nameable (overhangs, low bases, row-filling surfaces).

\keyline{Applications.} One passive camera and a few-watt accelerator suit small unmanned ground vehicles where cost,
power and size are limited, notably in the defence industry, where a camera also has the advantage of emitting nothing,
unlike LiDAR or active depth sensors, that could reveal the vehicle.

\keyline{An evaluation that explains itself.} Oracle, exact-obstacle, simulator-pose and own-pose arms decompose the gap to a
perfect score into execution (0.6), perception (1.3) and localisation (0.2 clean, within seed noise; 0.9 reached), so \cref{tab:negative}'s
interventions were not tuned blindly.

\section{Limitations}
The suite is one reconstructed yard with static obstacles; the wheel encoder and gyro the own-pose results rely on are simulated, and on the real drive the onboard SLAM, tuned in
simulation, beats dead reckoning from the commands ($-0.11$ vs $-0.31$) but not the vision SLAM baselines;
the student's metric range on real ground is not calibrated and the camera pitch is measured (3.8$^\circ$) but not applied onboard; and the Pi's camera path, power draw and
an autonomous drive of the real car remain to be measured (Section~\ref{sec:real} is a manual drive). The real results come from one drive, one route and one lighting condition, without ground truth. Obstacle bases below 5\,cm are
invisible to the scan, and the planner commits to routes through space the camera has not seen. Absolute HITL scores are specific to the simulator build (a restored build scores the released code 1.4 goals per seed lower); all comparisons were within one build.

\section{Conclusion}
A foundation-quality monocular depth network can run on a \rpi{} with a \hailo{}, but not as a transformer. Distilled into an
attention-free student it runs at 27~FPS, its int8 output equals its float output, and on held-out scenes of its simulated yard
it is the most accurate depth network on the chip; on stereo-verified real outdoor points it is not resolvably different from the vendor's
\dav-Small. Ranged by ground-contact geometry, its relative depth becomes a metric scan without metric training, and the Pi itself
maps, plans, follows and localises on it. In a real-time hardware-in-the-loop run through a reconstructed yard, the system reaches the goal without
contact in 7.9$\pm$1.1 of 10 scenes with simulator pose, 7.7$\pm$0.7 on its own pose and 8.2$\pm$0.4 with the real board in the loop (five seeds), where the same stack
with perfect obstacles reaches 9.2$\pm$0.4. The measured gap points at two concrete next steps: perception of low structure and
exploration that looks before it commits.

\section*{Acknowledgment}
We thank the Department of Computer Science at Ariel University for its support.

\bibliographystyle{IEEEtran}
\bibliography{refs}
\end{document}

%% file: tables/chipfid.tex
\newcommand{\chipfidpfifty}{0.99995}\newcommand{\chipfidpfive}{0.99897}
\newcommand{\chipfidvfivepfifty}{0.99993}\newcommand{\chipfidvfivepfive}{0.99569}

%% file: tables/paper_bench.tex
\begin{tabular}{@{}lrrrrr@{}}
\toprule
Model & \multicolumn{2}{c}{AbsRel$\downarrow$} & $\delta_1\uparrow$ & $r\uparrow$ & Speed \\
\cmidrule(lr){2-3}
 & all & obst. & all & real & \\
\midrule
\multicolumn{6}{@{}l}{\emph{On the Hailo-8L (int8), FPS on the Pi}} \\
\textbf{Ours, student v5} & 0.269 & \textbf{0.197} & 0.669 & 0.996 & 27.1 \\
Ours, student v3 & \textbf{0.269} & 0.204 & 0.674 & 0.995 & 27.1 \\
DAv2-S, Hailo zoo & 0.354 & 0.334 & 0.593 & 0.979 & 21.6 \\
DAv2-S, our int8 & 0.477 & 0.330 & 0.423 & 0.982 & 16.6 \\
SC-DepthV3, zoo & 0.414 & 0.352 & 0.506 & 0.950 & 53.6 \\
FastDepth, zoo & 0.530 & 0.767 & 0.344 & 0.709 & 121.3 \\
\midrule
\multicolumn{6}{@{}l}{\emph{Float on the RTX 5090 GPU, ms per frame}} \\
DAv2-L (teacher) & 0.284 & 0.144 & 0.643 & 1.000 & 175\,ms \\
DAv2-S & 0.283 & 0.150 & 0.651 & 0.999 & 39\,ms \\
DA3-Small & 0.315 & 0.172 & 0.611 & 0.988 & 42\,ms \\
DA3-Base & 0.304 & 0.188 & 0.616 & 0.985 & 40\,ms \\
\bottomrule
\end{tabular}

%% file: tables/paper_hitl.tex
\begin{tabular}{@{}lrrr@{}}
\toprule
Configuration & $n$ & Clean & Reached \\
\midrule
Oracle: exact props, lockstep & 5 & 9.8 $\pm$ 0.4 & 10.0 \\
Real time, exact obstacles & 5 & 9.2 $\pm$ 0.4 & 9.8 \\
\midrule
Student v3, GT pose & 10 & 7.9 $\pm$ 1.0 & 9.0 \\
\textbf{Student v5, GT pose} & 10 & 7.9 $\pm$ 1.1 & 9.0 \\
\midrule
Student v3, own pose & 10 & 7.1 $\pm$ 0.9 & 7.4 \\
\textbf{Student v5, own pose} & 10 & 7.7 $\pm$ 0.7 & 8.1 \\
\textbf{v5, own pose, real Pi + Hailo} & 5 & 8.2 $\pm$ 0.4 & 8.4 \\
Camera-only VO (v1, earlier planner) & 5 & 2.6 $\pm$ 0.9 & 3.6 \\
\bottomrule
\end{tabular}

%% file: paper.bbl
\begin{thebibliography}{10}
\providecommand{\url}[1]{#1}
\csname url@samestyle\endcsname
\providecommand{\newblock}{\relax}
\providecommand{\bibinfo}[2]{#2}
\providecommand{\BIBentrySTDinterwordspacing}{\spaceskip=0pt\relax}
\providecommand{\BIBentryALTinterwordstretchfactor}{4}
\providecommand{\BIBentryALTinterwordspacing}{\spaceskip=\fontdimen2\font plus
\BIBentryALTinterwordstretchfactor\fontdimen3\font minus
  \fontdimen4\font\relax}
\providecommand{\BIBforeignlanguage}[2]{{%
\expandafter\ifx\csname l@#1\endcsname\relax
\typeout{** WARNING: IEEEtran.bst: No hyphenation pattern has been}%
\typeout{** loaded for the language `#1'. Using the pattern for}%
\typeout{** the default language instead.}%
\else
\language=\csname l@#1\endcsname
\fi
#2}}
\providecommand{\BIBdecl}{\relax}
\BIBdecl

\bibitem{ranftl2022midas}
R.~Ranftl, K.~Lasinger, D.~Hafner, K.~Schindler, and V.~Koltun, ``Towards
  robust monocular depth estimation: Mixing datasets for zero-shot
  cross-dataset transfer,'' \emph{IEEE Transactions on Pattern Analysis and
  Machine Intelligence}, vol.~44, no.~3, pp. 1623--1637, 2022.

\bibitem{yang2024da1}
L.~Yang, B.~Kang, Z.~Huang, X.~Xu, J.~Feng, and H.~Zhao, ``Depth anything:
  Unleashing the power of large-scale unlabeled data,'' in \emph{IEEE/CVF
  Conference on Computer Vision and Pattern Recognition (CVPR)}, 2024.

\bibitem{yang2024dav2}
L.~Yang, B.~Kang, Z.~Huang, Z.~Zhao, X.~Xu, J.~Feng, and H.~Zhao, ``Depth
  anything v2,'' in \emph{Advances in Neural Information Processing Systems
  (NeurIPS)}, 2024.

\bibitem{bochkovskiy2025depthpro}
A.~Bochkovskiy, A.~Delaunoy, H.~Germain, M.~Santos, Y.~Zhou, S.~R. Richter, and
  V.~Koltun, ``Depth pro: Sharp monocular metric depth in less than a second,''
  in \emph{International Conference on Learning Representations (ICLR)}, 2025.

\bibitem{lin2025da3}
H.~Lin, S.~Chen, J.~H. Liew \emph{et~al.}, ``Depth anything 3: Recovering the
  visual space from any views,'' \emph{arXiv preprint arXiv:2511.10647}, 2025.

\bibitem{rpi2024aikit}
{Raspberry Pi Ltd}, ``Raspberry pi ai kit: product brief,''
  \url{https://datasheets.raspberrypi.com/ai-kit/raspberry-pi-ai-kit-product-brief.pdf},
  2024, hailo-8L M.2 module, 13 TOPS, published June 2024.

\bibitem{ranftl2021dpt}
R.~Ranftl, A.~Bochkovskiy, and V.~Koltun, ``Vision transformers for dense
  prediction,'' in \emph{IEEE/CVF International Conference on Computer Vision
  (ICCV)}, 2021.

\bibitem{hu2024metric3dv2}
M.~Hu, W.~Yin \emph{et~al.}, ``Metric3d v2: A versatile monocular geometric
  foundation model for zero-shot metric depth and surface normal estimation,''
  \emph{IEEE Transactions on Pattern Analysis and Machine Intelligence}, 2024.

\bibitem{piccinelli2024unidepth}
L.~Piccinelli, Y.-H. Yang, C.~Sakaridis, M.~Segu, S.~Li, L.~Van~Gool, and
  F.~Yu, ``Unidepth: Universal monocular metric depth estimation,'' in
  \emph{IEEE/CVF Conference on Computer Vision and Pattern Recognition (CVPR)},
  2024.

\bibitem{he2025distillanydepth}
X.~He, D.~Guo, H.~Li, R.~Li, Y.~Cui, and C.~Zhang, ``Distill any depth:
  Distillation creates a stronger monocular depth estimator,'' \emph{arXiv
  preprint arXiv:2502.19204}, 2025.

\bibitem{wofk2019fastdepth}
D.~Wofk, F.~Ma, T.-J. Yang, S.~Karaman, and V.~Sze, ``Fastdepth: Fast monocular
  depth estimation on embedded systems,'' in \emph{IEEE International
  Conference on Robotics and Automation (ICRA)}, 2019.

\bibitem{zhang2023litemono}
N.~Zhang, F.~Nex, G.~Vosselman, and N.~Kerle, ``Lite-mono: A lightweight cnn
  and transformer architecture for self-supervised monocular depth
  estimation,'' in \emph{IEEE/CVF Conference on Computer Vision and Pattern
  Recognition (CVPR)}, 2023.

\bibitem{sun2024scdepthv3}
L.~Sun, J.-W. Bian \emph{et~al.}, ``Sc-depthv3: Robust self-supervised
  monocular depth estimation for dynamic scenes,'' \emph{IEEE Transactions on
  Pattern Analysis and Machine Intelligence}, vol.~46, no.~1, pp. 497--508,
  2024.

\bibitem{ma2026asyncmde}
L.~Ma, Y.~Li, B.~Jiang, Z.~Zhong, H.~Ding, and L.~Zhu, ``Asyncmde: Real-time
  monocular depth estimation via asynchronous spatial memory,'' \emph{arXiv
  preprint arXiv:2603.10438}, 2026.

\bibitem{lin2022fqvit}
Y.~Lin, T.~Zhang, P.~Sun, Z.~Li, and S.~Zhou, ``Fq-vit: Post-training
  quantization for fully quantized vision transformer,'' in \emph{International
  Joint Conference on Artificial Intelligence (IJCAI)}, 2022.

\bibitem{yuan2022ptq4vit}
Z.~Yuan, C.~Xue, Y.~Chen, Q.~Wu, and G.~Sun, ``Ptq4vit: Post-training
  quantization for vision transformers with twin uniform quantization,'' in
  \emph{European Conference on Computer Vision (ECCV)}, 2022.

\bibitem{li2023repqvit}
Z.~Li, J.~Xiao, L.~Yang, and Q.~Gu, ``Repq-vit: Scale reparameterization for
  post-training quantization of vision transformers,'' in \emph{IEEE/CVF
  International Conference on Computer Vision (ICCV)}, 2023.

\bibitem{simon2023mononav}
N.~Simon and A.~Majumdar, ``Mononav: Mav navigation via monocular depth
  estimation and reconstruction,'' in \emph{International Symposium on
  Experimental Robotics (ISER)}, 2023.

\bibitem{sharma2026monompc}
B.~Sharma, P.~Jadhav, P.~Paul, K.~M. Krishna, and A.~K. Singh, ``Monompc:
  Monocular vision based navigation with learned collision model and risk-aware
  model predictive control,'' \emph{IEEE Robotics and Automation Letters
  (arXiv:2508.07387)}, 2026.

\bibitem{finke2026omninav}
J.~Finke, W.~P. Martis, A.~Schmelter, L.~Erbach, C.~Jestel, and M.~Wiedemann,
  ``Learning vision-based omnidirectional navigation: A teacher-student
  approach using monocular depth estimation,'' \emph{arXiv preprint
  arXiv:2603.01999}, 2026.

\bibitem{shah2023vint}
D.~Shah, A.~Sridhar, N.~Dashora, K.~Stachowicz, K.~Black, N.~Hirose, and
  S.~Levine, ``Vint: A foundation model for visual navigation,'' in
  \emph{Conference on Robot Learning (CoRL)}, 2023.

\bibitem{sridhar2024nomad}
A.~Sridhar, D.~Shah, C.~Glossop, and S.~Levine, ``Nomad: Goal masked diffusion
  policies for navigation and exploration,'' in \emph{IEEE International
  Conference on Robotics and Automation (ICRA)}, 2024.

\bibitem{yang2023iplanner}
F.~Yang, C.~Wang, C.~Cadena, and M.~Hutter, ``iplanner: Imperative path
  planning,'' in \emph{Robotics: Science and Systems (RSS)}, 2023.

\bibitem{roth2024viplanner}
P.~Roth, J.~Nubert, F.~Yang, M.~Mittal, and M.~Hutter, ``Viplanner: Visual
  semantic imperative learning for local navigation,'' in \emph{IEEE
  International Conference on Robotics and Automation (ICRA)}, 2024.

\bibitem{dolgov2010hybrid}
D.~Dolgov, S.~Thrun, M.~Montemerlo, and J.~Diebel, ``Path planning for
  autonomous vehicles in unknown semi-structured environments,'' \emph{The
  International Journal of Robotics Research}, vol.~29, no.~5, pp. 485--501,
  2010.

\bibitem{coulter1992purepursuit}
R.~C. Coulter, ``Implementation of the pure pursuit path tracking algorithm,''
  Carnegie Mellon University, Robotics Institute, Tech. Rep. CMU-RI-TR-92-01,
  1992.

\bibitem{he2016resnet}
K.~He, X.~Zhang, S.~Ren, and J.~Sun, ``Deep residual learning for image
  recognition,'' in \emph{IEEE Conference on Computer Vision and Pattern
  Recognition (CVPR)}, 2016.

\bibitem{ronneberger2015unet}
O.~Ronneberger, P.~Fischer, and T.~Brox, ``U-net: Convolutional networks for
  biomedical image segmentation,'' in \emph{Medical Image Computing and
  Computer-Assisted Intervention (MICCAI)}, 2015.

\bibitem{lucas1981lk}
B.~D. Lucas and T.~Kanade, ``An iterative image registration technique with an
  application to stereo vision,'' in \emph{International Joint Conference on
  Artificial Intelligence (IJCAI)}, 1981.

\bibitem{nvidia2025isaaclab}
{NVIDIA} \emph{et~al.}, ``Isaac lab: A gpu-accelerated simulation framework for
  multi-modal robot learning,'' \emph{arXiv preprint arXiv:2511.04831}, 2025.

\bibitem{kerbl2023gs}
B.~Kerbl, G.~Kopanas, T.~Leimk{\"u}hler, and G.~Drettakis, ``3d gaussian
  splatting for real-time radiance field rendering,'' \emph{ACM Transactions on
  Graphics}, vol.~42, no.~4, 2023.

\bibitem{wu2025_3dgut}
Q.~Wu, J.~Martinez~Esturo, A.~Mirzaei, N.~Moenne-Loccoz, and Z.~Gojcic,
  ``3dgut: Enabling distorted cameras and secondary rays in gaussian
  splatting,'' in \emph{IEEE/CVF Conference on Computer Vision and Pattern
  Recognition (CVPR)}, 2025.

\end{thebibliography}
